\documentclass{article}

\usepackage{microtype}
\usepackage{graphicx}
\usepackage{subfigure}
\usepackage{booktabs} 
\usepackage{adjustbox}
\usepackage{wrapfig}
\usepackage{hyperref}
\usepackage{multirow}

\usepackage{amsmath}

\DeclareMathOperator*{\argmin}{arg\,min}

\usepackage{enumitem}
\setlist[itemize]{noitemsep, topsep=0pt}

\input{Definitions}
\usepackage[accepted]{icml2021}

\icmltitlerunning{Latent Neural Fokker-Planck Kernels}

\begin{document}

\twocolumn[
\icmltitle{Learning High-Dimensional Distributions with \\ Latent Neural Fokker-Planck Kernels}



\icmlsetsymbol{equal}{*}

\begin{icmlauthorlist}
\icmlauthor{Yufan Zhou}{to}
\icmlauthor{Changyou Chen}{to}
\icmlauthor{Jinhui Xu}{to}
\end{icmlauthorlist}

\icmlaffiliation{to}{Department of Computer Science and Engineering, State University of New York at Buffalo}

\icmlcorrespondingauthor{Yufan Zhou}{yufanzho@buffalo.edu}

\icmlkeywords{Machine Learning}

\vskip 0.3in
]



\printAffiliationsAndNotice{}  

\begin{abstract}

Learning high-dimensional distributions is an important yet challenging problem in machine learning with applications in various domains. In this paper, we introduce new techniques to formulate the problem as solving Fokker-Planck equation in a lower-dimensional latent space, aiming to mitigate challenges in high-dimensional data space. Our proposed model consists of latent-distribution morphing, a generator and a parameterized Fokker-Planck kernel function. One fascinating property of our model is that it can be trained with arbitrary steps of latent distribution morphing or even without morphing, which makes it flexible and as efficient as Generative Adversarial Networks (GANs). Furthermore, this property also makes our latent-distribution morphing an efficient plug-and-play scheme, thus can be used to improve arbitrary GANs, and more interestingly, can effectively correct failure cases of the GAN models. Extensive experiments illustrate the advantages of our proposed method over existing models.

\end{abstract}

\section{Introduction}
Learning a complex distribution from high-dimensional data is an important and challenging problem in machine learning with 
a variety of real-world applications
such as 
synthesizing high-fidelity data. 
Some representative approaches for solving this problem 
include generative adversarial networks (GANs) \citep{goodfellow2014generative}, variational auto-encoder (VAEs) \citep{Kingma2014}, normalizing flows \citep{rezende2015variational, kingma2018glow}, enegy-based models (EBMs) \cite{NIPS2019_8619}, {\it etc}. Despite their tremendous successes, each of them also suffers from its own intrinsic limitations which are partially caused by the complex nature of the high-dimensional data space.  


GAN \cite{goodfellow2014generative} is perhaps one of the most popular generative models due to its ability to generate high-fidelity data. 
In recent years, there has been a surge of interest in developing various types of GANs  
by designing new objectives \cite{li2017mmd, arjovsky2017wasserstein, genevay2018learning, wang2018improving} or stabilizing the training process \cite{miyato2018spectral, gulrajani2017improved, arbel2018gradient}. 
A common and well-known issue of all such GANs is the seemingly unavoidable mode collapse. We 
conjecture 
that this is partially due to the following two reasons: 1) GAN directly optimizes its objective function in the often complicated high-dimensional data space, and thus is more likely to get stuck at 
some sub-optimal solutions; and 2) GAN aims to learn a mapping from a flat and single-mode latent distribution ({\it e.g.}, Gaussian) to a complex, high dimensional and possibly multi-mode data distribution. This complicated mapping could impose 
a tremendous challenge to learn all modes
in the optimization 
process.

VAE and Normalizing Flow based methods are likelihood-driven models, making them more tractable but potentially less expressive compared to GANs. A consequence is that 
the generated data from these models are smoother and less real-looking. 
EBM assumes that the data distribution follows an unnormalized energy-based form, making it more tractable than GANs and more flexible than VAEs. However, its training requires samples generated from the model distribution, which is typically done via Langevin dynamics \citep{roberts1996exponential, welling2011bayesian}, which could be time-consuming in practice and could suffer considerably from the high domensionality of the data space. Recently, some works have been proposed to alleviate this problem, by combining a generator with the EBM \citep{kumar2019maximum, grathwohl2020nomcmc, Xie2020CooperativeTO, Arbel2020KALEWE}, we will discuss the major difference between our proposed method with theirs and provide experimental comparisons in later sections.

To deal with the aforementioned issues, we propose, in this paper, a new framework to learn 
the implicit distribution of high dimensional data via deep generative models (DGMs).
Our framework aims to achieve the following goals: 1) define the problem in the space of probability measure and avoid learning the complicated and possibly single-to-multi-mode matching from the latent space to the data space as in GAN; and 2) relieve from the requirement of sampling from data space for learning as in EBM. To realize these goals,
\begin{itemize}
    \item we formulate the problem of learning high-dimensional distribution as one that learns a neural Fokker-Planck kernel, inspired by the Fokker-Planck (FP) equation \citep{risken1996fokker};
    \item we define the Fokker-Planck kernel in a much lower dimensional latent space so that learning becomes morphing a flat latent distribution into a more complex one in the same latent space. Consequently, mapping from the latent space to the data space is carried out by
    a multi-mode distribution matching scheme, implemented by a parameterized generator.
\end{itemize}

\begin{figure}[t!]
    \centering
    \includegraphics[width=1.0\linewidth]{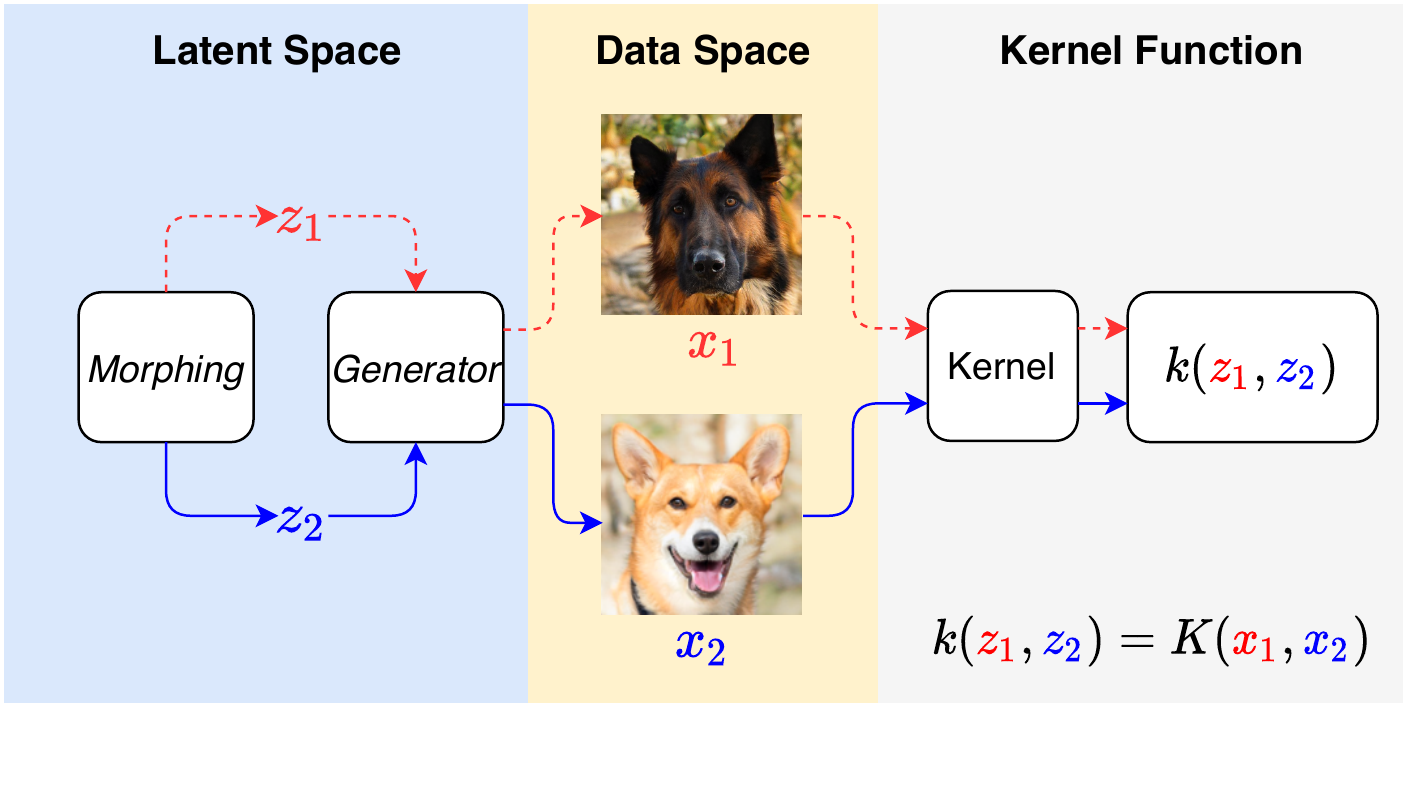}
    \vskip -0.25in
    \caption{An overview of the proposed framework. The key components include a latent-distribution morphing mechanism, a parameterized generator and a neural Fokker-Planck kernel.}
    \label{fig:model_figure}
\end{figure}

In summary, our framework consists of three components: 1) a parameterized kernel function representing the FP kernel; 2) a latent-distribution morphing scheme implemented via gradient flows in the latent space; and 3) a generator mapping samples in the latent space to samples in the data space. The connection between the three components is illustrated in Figure \ref{fig:model_figure}. A remarkable feature of our framework is that the latent-distribution morphing scheme allows us to develop a plug-and-play mechanism directly from pre-trained GAN models, where we use our latent-distribution morphing scheme to optimize their flat latent distributions. This gives us a fine-tuning mechanism independent of the original GAN models, which can lead to significant improvement over the original GANs. In experiments, we show that our latent-distribution morphing scheme can be used to mitigate the mode collapses problem, indicating that mode collapse is indeed partially caused by the mis-matching between the simple latent distribution and the complex data distribution. 
Moreover, based on a number of state-of-the-art architecture settings and extensive experiments,
we demonstrate 
the advantages of our proposed framework over existing models on a number of tasks, including high-fidelity image generation and image translation.


\section{The Proposed Framework}
In this paper, we follow the commonly adopted assumption that the observed data are sampled from a lower-dimensional manifold embedded in the high-dimensional space. Let $\mathcal{Z}$ be the 
low-dimensional latent space\footnote{We should not confuse $\mathcal{Z}$ with the underlying manifold.}, $\mathcal{X}$ be the data space, and $g: \mathcal{Z} \rightarrow \mathcal{X}$ be a mapping from the latent representations to data samples. Let $p(\zb), q(\zb), \zb \in \mathcal{Z}$, be two distributions in the latent space, and $P(\xb)$ and $Q(\xb)$, $\xb \in \mathcal{X}$\footnote{With a little abuse of notation for the purpose of conciseness, we will not distinguish the probability density functions and probability measures represented by $p$, $q$, $P$, and $Q$.},  be their corresponding distributions in the data space induced by the mapping
$g$ ($g$ is also called the generator).  

Assume that $P$ is the target distribution to be learned from a set of observed data. Instead of directly learning $P$, we propose to learn its associated latent distribution $p$ first. To this end, by the variational principle, we will learn a distribution $q$ such that it is close to $p$ under some metric. A natural way to obtain $q$ is to formulate the problem in $\mathcal{P}_2$, the space of probability measures endowed with the 2-Wasserstein distance \cite{santambrogio2016}, as:
\begin{align}\label{eq:optimization_w2}
    q^*  = \arg\min_{q\in\mathcal{P}_2} F(q, p)~,
\end{align}
where $F(q, p)$ is a functional measuring the difference between $p$ and $q$. 
A popular choice of $F$ is the KL-divergence, which would make \eqref{eq:optimization_w2} equivalent to maximizing the negative log-likelihood $-\mathbb{E}_{p(\zb)}\left[ \log (q(\zb))\right]$. For a more general setting, we instead focus on the $f$-divergence $D_f(p_D \Vert q)$, defined as 
\[
D_{f}(p \Vert q) = \int_{\mathcal{Z}} q(\zb) f(p(\zb)/q(\zb)) \text{d}\zb~,
\]
where $f: (0, \infty) \rightarrow R$ is a convex function with $f(1) = 0$. Many distribution metrics are special cases of $f$-divergence, including the Kullback-Leibler (KL) divergence and the Jensen-Shannon (JS) divergence. An important property of $f$-divergence is its convexity, {\it i.e.}, $p \rightarrow D_{f}(p \Vert q)$ and $q \rightarrow D_{f}(p \Vert q)$ are both convex. 

\paragraph{Default solutions via Wasserstein gradient flows (WGFs)}
With the convexity of the $f$-divergence, a default solution to problem \eqref{eq:optimization_w2} is by applying the method of WGF \citep{ambrosio2008gradient, santambrogio2016}. Specifically, \eqref{eq:optimization_w2} is re-written as a WGF in the space of probability measures, which is represented by a partial differential equation (PDE):
\begin{align}\label{eq:optimization_f_divergence}
    \dfrac{\partial q}{\partial t} = - \nabla _{W_2} F(q, p) \triangleq - \nabla _{W_2} D_{f}(p\Vert q)~,
\end{align}
The following lemma ensures the convergence of the $f$-divergence under  WGF.
\begin{lemma}\label{thm:convergence}
Assume that $(q^t)_{t\geq 0}$ evolves according to the WGF of $f$-divergence, $F(q^t, p) = D_{f}(p \Vert q^t)$, with initialization $q^0$. Let $d_{W_2}(\cdot, \cdot)$ be the 2-Wasserstein distance. Assume that $F(q^0, p)<\infty$ and $d_{W_2}(q^0, p)<\infty$. Then, the following holds:\\
1) $F(q^t, p)$ is non-increasing and converges to the global optimum as $t\rightarrow \infty$;\\
2) If $\Vert \nabla_{W_2} F(q^t, p) \Vert ^2 \geq \lambda F(q^t, p)$ with some constant $\lambda >0$, then $F(q^t, p) \leq \exp(-\lambda t) F(q^0, p)$;\\
3) If $F(q^t, p)$ is $\lambda$-geodesically convex\footnote{See Appendix for the definition of $\lambda$-geodesically convex.} with some constant $\lambda>0$, then $d_{W_2}(q^t, p) \leq \exp(-\lambda t)d_{W_2}(q^0, p)$;\\
4) If mapping $g$ is Lipschitz continuous, then $d_{W_2}(Q^t, P)$ converges with the same convergence rate as $d_{W_2}(q^t, p)$.
\end{lemma}
To solve the WGF problem \eqref{eq:optimization_f_divergence}, a popular way is to use the  Jordan-Kinderlehrer-Otto (JKO) scheme \cite{jordan1998variational}, which derive the solution of the PDE via solving a sequence of optimization problems: $q^{k+1}(\zb) = \argmin _{q} D_{f}(p \Vert q) + d^2_{W_2}(q^k, q)/(2h)$, 
where $k=0,1,2...$, $h$ is a hyper-parameter. In practice, $d^2_W$ is computationally infeasible, and thus can be approximated with different methods \citep{cuturi2013sinkhorn, chizat2020faster}. However, a potential issue of previous works using this method is that optimization is performed directly on the high-dimensional data space, which could be severely impacted by the curse of dimensionality.

\subsection{Reformulation with Latent Neural FP Kernels}
It is known that $ \nabla _{W_2} F(q, p)  = - div (q \nabla_{\zb} \delta F/\delta q)$ \citep{santambrogio2016}, where $\delta F/\delta q$ denotes the first variation of functional $F$ at $q$, and $div$ denotes the divergence operator. With this, we can re-write \eqref{eq:optimization_f_divergence} as:
\begin{align}\label{eq:optimization_fp_equation}
    \dfrac{\partial q}{\partial t} = div (q \nabla_{\zb} \psi (\zb)), \text{where } \psi (\zb) = \dfrac{\delta D_{f}(p \Vert q)}{\delta q}~.
\end{align}
\eqref{eq:optimization_fp_equation} is a special case of the well-known FP equation. In the following, we propose a kernel based parameterization inspired by the FP kernel \citep{risken1996fokker,carrillo1998exponential, Bilal_2020} to solve \eqref{eq:optimization_fp_equation}. One advantage of the FP kernel is that it can be solved efficiently with simple and mesh-free methods, which can be applied on irregular and complicated spaces. 

Following \cite{Bilal_2020}, we denote the FP kernel as $k(t, \xb, s, \yb)$, with $t$ and $s$ being the time indexes. 
For the FP equation \eqref{eq:optimization_fp_equation}, the theory of FP kernels reveals that the density function $q^t(\zb)$ at any time $t$ is characterized by a convolutional form as \cite{Bilal_2020}:
\begin{align}\label{eq:fpkernel}
    q^t(\zb) = \int_{\mathcal{Z}} k(t, \zb, s, \zb^\prime) q^s(\zb^\prime) \exp(V^s(\zb^\prime)) \text{d}\zb^\prime~,
\end{align}
where $V$ is an arbitrary priori function, and $s$ indexes arbitrary time. 
According to Lemma \ref{thm:convergence}, we expect $q(\zb)$ to converge to $p(\zb)$ at the infinite-time limit, {\it i.e.}, $q^{\infty}(\zb) = p(\zb)$. Consequently, taking $s\rightarrow \infty$ in \eqref{eq:fpkernel}, our implicit latent distribution $q(\zb)$ can be written as
\begin{align*}
    q(\zb) = \dfrac{\int_{\mathcal{Z}} k(\zb, \zb^\prime) p(\zb^\prime) \text{d}\zb^\prime}{Z} \triangleq  \dfrac{\tilde{q}( \zb)}{Z}, 
\end{align*}
where $Z = \int \tilde{q}(\xb) d \xb$ is the normalizer, and 
the $V$ function in \eqref{eq:fpkernel} has been defined as  
a special constant function. Furthermore, we 
will not directly describe the dependency of $k$ on $t$ in the following as it corresponds to the optimization process. Due to the fact that we have access only to the real samples in the data space but not in the latent space, we propose to construct a deep kernel via the generator $g$ as $k = K\circ g$ with $K$ being a kernel function in the data space:
\begin{align}\label{eq:composite_kernl}
    k(\zb, \zb^\prime) = K(g(\zb), g(\zb^\prime)) = K(g(\zb), \xb^\prime)~,
\end{align}
where $\xb^\prime$ corresponds to a training data sample from $P(\xb^\prime)$. We will construct $K$ to be a deep kernel (see \eqref{eq:kernel} below) to mitigate challenges in the high-dimensional data space. With such as kernel construction, $q(\zb)$ can be written as
\begin{align}\label{eq:density_parameterization}
q(\zb) = \dfrac{\mathbb{E}_{P(\xb^\prime)}\left[ K(g(\zb), \xb^\prime)\right]}{Z}~.
\end{align}

\begin{remark}\label{re:ebm}
Our parameterization \eqref{eq:density_parameterization} is related to EBM \cite{NIPS2019_8619}. 
To see this, re-write our parameterization as $q(\zb) \propto \exp(\log\mathbb{E}_{P( \xb^\prime) } \left[K(g(\zb),  \xb^\prime)\right])$. 
An EBM defined in the latent space with the form $q(\zb) \propto \exp(-E(\zb))$  is recovered if the energy function $E$ is defined as $E(\zb) \triangleq -\log\mathbb{E}_{P( \xb^\prime) } \left[K(g(\zb),  \xb^\prime)\right]$ in the latent space.
\end{remark}

In practice, we parameterize $k$ and $g$ with neural networks, written as $f_{\phib}$ and $g_{\thetab}$ with parameters $\phib$ and $\thetab$, respectively. This leads to re-writing \eqref{eq:composite_kernl} as
\begin{align} \label{eq:kernel}
\hspace{-0.3cm}k_{\phib, \thetab}(\zb, \zb^\prime) = K_{\phib}(\xb,  \xb^\prime) = \exp(-\Vert f_{\phib}(\xb) - f_{\phib}( \xb^\prime)\Vert^2),
\end{align}
where $\zb \sim q(\zb), \zb^\prime \sim p(\zb^\prime), \xb =g_{\thetab}(\zb)$, and  $\xb^\prime$ is the data sample corresponding to latent code $\zb^\prime$. Note that $K$ and $k$ are valid, positive semi-definite kernels \citep{shawe2004kernel,li2017mmd} by construction. There are other methods to construct potentially more expressive kernels \citep{rahimi2007random, samo2015generalized, remes2017non, li2019implicit, zhou2019kernelnet}. For simiplicity, we only consider the parameterization in \eqref{eq:kernel}.

\subsection{Latent-Distribution Morphing}
We now discuss how to generate latent codes following $q(\zb)$. Starting from a simple latent distribution, we define a new WGF in the latent space whose stationary solution is $q(\zb)$. With this, morphing the latent distribution corresponds to solving the WGF to make it converge to $q(\zb)$. We implement this by deriving particle update rules for each latent code $\zb$.

Specifically, let $r(\zb)$ be the distribution to be updated, implemented by minimizing the $f$-divergence $D_f(r \Vert q)$. This corresponds to the following WGF:
\begin{align}\label{eq:latentwgf}
    \dfrac{\partial r}{ \partial t} = - \nabla_{W_2} D_{f}( r\Vert q) = div (r \nabla_{\zb} \dfrac{\delta D_f(r \Vert q)}{\delta r} )~.
\end{align}
According to \cite{ito1951stochastic, risken1996fokker}, the corresponding particle optimization can be described by the following ordinary differential equation (ODE):
\begin{align}\label{eq:zode}
    \text{d} \zb = - \nabla_{\zb} \dfrac{\delta D_f(r \Vert q)}{\delta r} \text{d}t~.
\end{align}
Note that this WGF \eqref{eq:latentwgf} is different from \eqref{eq:optimization_fp_equation}, which has a different "target distribution" $q(\zb)$ instead of $p(\zb)$. 
We propose to use kernel density estimator \citep{10.5555/1202956} to estimate $r(\zb)$, {\it i.e.}, $r(\zb) = \mathbb{E}_{r(\zb^\prime)}\left[K_{\phib}(g_{\thetab}(\zb), g_{\thetab}(\zb^\prime))\right] \approx \sum_{i=1}^n K_{\phib}(g_{\thetab}(\zb), g_{\thetab}(\zb^\prime_i))/n$. We solve $\zb$ iteratively by substituting the above $r(\zb)$ into  ODE \eqref{eq:zode}. Using superscript $m$ to index the iteration number and solving \eqref{eq:zode} result in
\begin{align}\label{eq:latent_sampling}
    \zb^{m + 1} = \zb^m - \lambda \nabla_{\zb}  \dfrac{\delta D_f(r \Vert q )}{\delta r}, 
\end{align}
where $\lambda$ is the step size; and $\zb^0 \sim \text{Uniform} \left[-1, 1\right]^{dim(\zb)}$.
We consider different forms of  $D_f$, including reverse Kullback-Leibler (RKL), Jensen-Shannon (JS), and Squared Hellinger (SH) divergence. Table \ref{tab:different_update_rule} provides different update equations for $\zb$. Detailed derivations are provided in the Appendix.

\begin{remark}
    Our kernel representation can also be applied to the standard Langevin sampling \citep{roberts1996exponential, welling2011bayesian} to derive an update equation for $\zb$. Recall that $q(\zb) \propto \mathbb{E}_{P(\xb^\prime)}\left[ K(g(\zb), \xb^\prime)\right]$. Given $\{\xb^\prime_i\}_{i=1}^n$ from $P(\xb^\prime)$, we can simply apply Langevin sampling based on the above $q(\zb)$, resulting in 
    \begin{align}\label{eq:langevin_sampling}
    \zb^{m + 1} = \zb^m + \lambda \nabla_{\zb} \log \sum_{i=1}^n K_{\phib}(g_{\thetab}(\zb^m),  \xb^\prime_i) + \epsilonb^m,
    \end{align}
    where $\epsilonb^m \sim \mathcal{N}(0, 2\lambda)$. Compared to the formulas in Table \ref{tab:different_update_rule}, our method represents an interacting particle system that considers the relationships between $\{\zb_i\}_{i=1}^n$. Specifically, $\sum_{i=1}^n K_{\phib}(g_{\thetab}(\zb), g_{\thetab}(\zb_i))$ is jointly minimized w.r.t. to all $\zb_i$'s during the sampling, leading to diverse representations. Performance comparisons with different update equations in Table~\ref{tab:different_update_rule} are provided in the experiment section, our proposed morphing \eqref{eq:latent_sampling} is shown to outperform the standard Langevin sampling method in \eqref{eq:langevin_sampling}.
\end{remark}

\begin{table}[t!]
    \caption{Different update gradient of different $f$-divergence}
    \label{tab:different_update_rule}
    \centering
    \begin{sc}
    \begin{adjustbox}{scale=0.75}
    \begin{tabular}{lc}
        \toprule
         Functional & Update Gradient $\nabla_{\zb} \dfrac{\delta D_f(r \Vert q )}{\delta r}$ \\
         \midrule 
         Langevin & $-\nabla_{\zb} \log \sum_i^n K(g(\zb), \xb^\prime_i)$\\
         KL & $-\nabla_{\zb} \log\sum_i^n K(g(\zb), \xb^\prime_i)+\nabla_{\zb}\log \sum_i^n K(g(\zb), g(\zb_i))$\\
         RKL & $- \nabla_{\zb} \left[  \sum_i^n K(g(\zb), \xb^\prime_i) / \sum_i^n K(g(\zb), g(\zb_i))  \right]$\\
         SH & $- \nabla_{\zb} \sqrt{  \sum_i^n K(g(\zb), \xb^\prime_i) / \sum_i^n K(g(\zb), g(\zb_i)) }$\\
          \multirow{2}{*}{JS} & $-\dfrac{1}{2}\{\nabla_{\xb}\log \left[ \sum_i^n K(g(\zb), g(\zb_i)) + \sum_i^n K(g(\zb), \xb^\prime_i) \right]$\\
          &$- \nabla_{\zb} \log 2\sum_i^n K(g(\zb), g(\zb_i))\}$\\
         \bottomrule
    \end{tabular}
    \end{adjustbox}
    \end{sc}
\end{table}

\subsection{Parameter Optimization}

Our optimization procedure aims to learn the parameters $(\phib, \thetab)$ for the kernel and the generator. We first derive update equations under the assumption that the latent distribution is globally optimized, {\it i.e.}, $r$ converges to $q$ in \eqref{eq:latentwgf}. We then consider the case where no latent-distribution morphing is implemented, combining with which we derive our solution for the case when $r$ does not fully converge to $q$.

\paragraph{Optimization under a globally-optimal latent distribution} 
In this case, latent samples are assumed to follow $q(\zb)$. Our goal is then to minimize the difference between $q(\zb)$ and $p(\zb)$, which requires computation of the gradients of $f$-divergence w.r.t the kernel parameters $\phib$ and the generator parameter $\thetab$. 

To derive gradient formulas for an arbitrary $f$-divergence, we will use its variational form \cite{nguyen2010estimating}:
$
D_{f}(p \Vert q) = \sup _{v \in \mathcal{V}} \mathbb{E}_{p(\zb)}\left[ v(\zb) \right] - \mathbb{E}_{q(\zb)}\left[ f^*(v(\zb))\right]
$, 
where $f^*$ is the Fenchel conjugate function of $f$ defined as $f^*(v) = \sup_{u \in R} uv - f(u)$; $\mathcal{V}$ is the function space such that an element of $\mathcal{V}$ is contained in the sub-differential of $f$ at $p/q$. Let $v^* \in \mathcal{V}$ be the function when the supremum is attained. A discussion on how to compute $f^*$ and $v^*$ is provided in the Appendix. Based on the variational form, we have the following result.

\begin{theorem}\label{thm:grad_of_f_divergence}
With $q(\zb)$ parameterized by \eqref{eq:density_parameterization}, the gradient of arbitrary $f$-divergence 
$
D_{f}(p \Vert q) = \mathbb{E}_{q(\zb)}\left[ f(p(\zb)/q(\zb))\right]
$
w.r.t. $\phib$ can be written as:
\begin{align}\label{eq:grad_of_f_divergence}
    &\nabla_{\phib}D_{f}(p \Vert q) \\
    = & -\mathbb{E}_{q(\zb)}\Big \{f^*(v^*(\zb)) \nabla_{\phib}\log\{ \mathbb{E}_{P(\xb^\prime)}\left[ K_{\phib}(g_{\thetab}(\zb), \xb^\prime)\right]\} \Big\} \nonumber \\
    + & \mathbb{E}_{q(\zb)}\left[ f^*(v^*(\zb))\right] \mathbb{E}_{q(\zb)}\big \{\nabla_{\phib} \log\{\mathbb{E}_{P(\xb^\prime)} \left[ K_{\phib}(g_{\thetab}(\zb), \xb^\prime)\right]\}\big \}. \nonumber 
\end{align}
In particular, for the KL divergence $D_{KL}(p \Vert q)$, the gradient w.r.t. $\phib$ is:
\begin{align}\label{eq:grad_of_kl}
    \nabla_{\phib}&D_{KL}(p \Vert q) =  - \mathbb{E}_{P(\xb)}\{\nabla_{\phib} \log\{\mathbb{E}_{P(\xb^\prime)} \left[ K_{\phib}(\xb, \xb^\prime)\right]\}\} \nonumber \\
&+ \mathbb{E}_{q(\zb)}\{\nabla_{\phib} \log\{\mathbb{E}_{P(\xb^\prime)} \left[ K_{\phib}(g_{\thetab}(\zb), \xb^\prime)\right]\}\}~.
\end{align}
\end{theorem}


\paragraph{Optimization without latent-distribution morphing}
In this case, we will have a fixed prior distribution for $\zb$, denoted as $r(\zb)$. 
In practice, as in GAN, we usually define $r(\zb)$ as a uniform distribution: $\text{Uniform} \left[-1, 1\right]^{dim(\zb)}$. With a set of samples $\{\zb_i\}_{i=1}^n$ from $r$,
we propose the following objective function to optimize the kernel:
\begin{align*}
    \mathcal{L}_{kde} = &  \dfrac{\sum_{j=1}^n}{n}\Big\Vert \tilde{r}(\zb_j) - \dfrac{1}{n} \Big\Vert^2 + \alpha \dfrac{\sum_{i,j=1}^n}{n^2}  K_{\phib}(g_{\thetab}(\zb_j), \xb^\prime_i), 
\end{align*}
where $\tilde{r}(\zb_j) \triangleq \sum_{i\neq j} K_{\phib}(g_{\thetab}(\zb_j), g_{\thetab}(\zb_i))/(n-1)$ is the kernel density estimator, which is expected to be close to the uniform distribution (the first term);  
similar to adversarial learning, the second term regularizes the kernel to distinguish generated samples with real samples. 
Interestingly, we find that only performing latent-distribution morphing in testing is usually good enough to improve the quality of the generated images.

\paragraph{Optimization under a potentially sub-optimal latent distribution}
Due to the numerical errors and limited iterations to solve \eqref{eq:zode} in practice, $r(\zb)$ is likely to converge to a sub-optimum. Since this can be considered as the case between the previous two cases, we propose to combine the two objective functions to learn the latent neural FP kernel:
\begin{align}\label{eq:grad_final_kernel}
    \nabla_{\phib} \mathcal{L} \triangleq \nabla_{\phib} \mathcal{L}_{kde} + \beta \nabla_{\phib}D_{f}(p \Vert r)~,
\end{align}
where $\beta$ is a hyper-parameter. 

\paragraph{Optimization for the generator}
With the learned latent neural FP kernel, we then use the maximum mean discrepancy (MMD) to optimize the generator, which essentially minimizes the difference between $P(\xb)$ and $R(\xb)$ \footnote{$R(\xb) = Q(\xb)$ in the case of a globally-optimal latent distribution. Although we train the kernel density estimator to estimate $r(\zb)$, the kernel may not estimate $R(\xb)$ well, because $R(\xb) = R(g_{\thetab}(\zb)) \neq r(\zb)$. However, if we assume $R(\xb) \propto r(\zb)$, we can actually train the generator by minimizing the negative log-likelihood instead of MMD. These two objectives lead to similar gradients, which is discussed in the Appendix.}.

\subsection{Plug-and-Play on Pre-trained Models}

\paragraph{Extension to arbitrary GAN} We describe how to apply our latent-distribution morphing to arbitrary GAN model, independent of training objectives and model architectures. Given a pre-trained GAN with generator $g_{\thetab}(\zb)$ and discriminator $d_{\phib}(\xb)$, we can simply construct a kernel using $g_{\thetab}(\zb)$ and  $d_{\phib}(\xb)$, 
\[
K_{\phib}(\xb, \yb) = \exp(-\Vert d_{\phib}(\xb) - d_{\phib}(\yb) \Vert^2)~,
\]
and only perform latent-distribution morphing via \eqref{eq:latent_sampling} during testing. The kernel in this case can be regarded as a sub-optimal solution to the Fokker-Planck equation. Our experiment results indicate that all GANs can be improved with the latent-distribution morphing. In particular, we show that most failure cases of GANs can be recovered, indicating that mode collapse can be caused by mis-matching between the latent space and the data space, and better generation can be achieved by learning a better matching between the two spaces. Furthermore, we also find that adding $\mathcal{L}_{kde}$ as a regularizer to train or fine-tune arbitrary GAN can also lead to performance improvement.
\paragraph{Extension to EBMs} 
Our proposed method can also be applied to construct new EBMs, including standard EBM \citep{NIPS2019_8619} and its variants \citep{yu2020training}. From Remark~\ref{re:ebm}, it is easy to see that if we remove the generator component in our framework, the model is essentially a kernel-based variant of the vanilla EBM, with the distribution morphing performing in the data space. If we keep the generator, then our method is related to some works on combining the generators with EBMs \citep{kumar2019maximum, Xie2020CooperativeTO, grathwohl2020nomcmc, Arbel2020KALEWE}. Different from these methods, sample energy in our model is evaluated by considering the similarities with the real samples via the kernel function. 
We have demonstrated in experiments that our kernel-based EBM obtains better performance than others.

\section{Comparison with Related Works}
\label{sec-rw}

Different from related works which try to combine generators with EBMs \citep{kumar2019maximum, Xie2020CooperativeTO, grathwohl2020nomcmc, Arbel2020KALEWE}, our proposed method not only achieves much better results on image generation tasks, it can also be extended to improve arbitrary pre-trained GAN models. Consequently, our method can be applied in many tasks where GAN-based models are applicable, e.g. image translation, while none of other works can be extended to these tasks.

Different from other techniques which tries to improve GANs by the idea of manipulating the latent distribution with signal from the discriminator \citep{tanaka2019discriminator, che2020your, ansari2020refining}, our propose method can significantly improve the performance of not only GANs, but also EBMs, while none of the aforementioned methods can be extended to EBMs. More importantly, most of previous works can only be applied to those GANs whose discriminator outputs scalar values. Although \cite{ansari2020refining} claims  that DG$f$low is capable of improving GANs with discriminators that output vector values, it requires an extra pre-trained discriminator and some fine-tuning, which could be inefficient. On the contrary, our proposed method can be directly applied to any pre-trained GANs, with no need of any extra network or fine-tuning.


\section{Experiments}
\subsection{Toy Experiments}

We first use toy experiments to demonstrate the effectiveness of our proposed framework in dealing with mode collapse, as well as to illustrate the latent-distribution morphing scheme. Our target is a weighted mixture of eight 2-D Gaussian distributions. The input (latent) noise to the generator are sampled from $\text{Uniform}\left[-1, 1 \right]$. All the networks consist of 2 fully-connected layers with 16 hidden units. 

\begin{figure}[t!]
    \centering
    \subfigure[Target]{\includegraphics[width=0.24\linewidth]{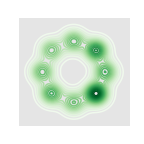}}
    \subfigure[GAN]{\includegraphics[width=0.24\linewidth]{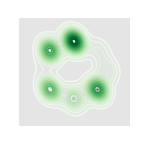}}  
    \subfigure[MMD]{\includegraphics[width=0.24\linewidth]{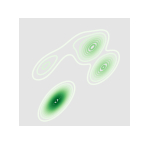}} 
    \subfigure[EBM-10]{\includegraphics[width=0.24\linewidth]{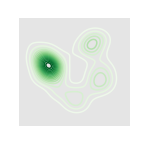}} 
    \subfigure[Ours]{\includegraphics[width=0.24\linewidth]{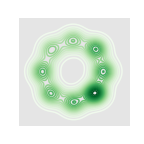}} 
    \subfigure[GAN-m]{\includegraphics[width=0.24\linewidth]{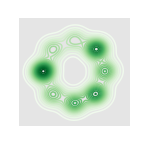}} 
    \subfigure[MMD-m]{\includegraphics[width=0.24\linewidth]{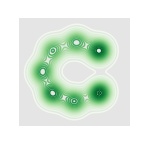}}   
    \subfigure[EBM-30]{\includegraphics[width=0.24\linewidth]{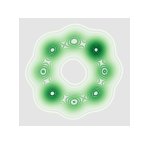}} 
    \caption{Results of toy experiments.
    EBM-$n$ ($n=10,30$) means EBM trained with $n$-step sampling; methods appended with ``-m'' mean that these methods are improved with latent morphing.}
    \label{fig:toy_exp}
\end{figure}

We compare our method with standard GAN, MMD-GAN, and EBM. We also test the plug-and-play functionality of our framework to improve the performance of standard GAN and MMD-GAN. Some results are provided in Figure \ref{fig:toy_exp}, from which we can see that our proposed model and EBM (trained with 30-step sampling) can recover all the modes, and our method learns a better approximation. Unexpectedly, GAN-based models fail to capture all the modes. However, after the plug-and-play refinement, they are able to capture all the modes. 

\begin{figure}
    \centering
    \includegraphics[width=0.7\linewidth]{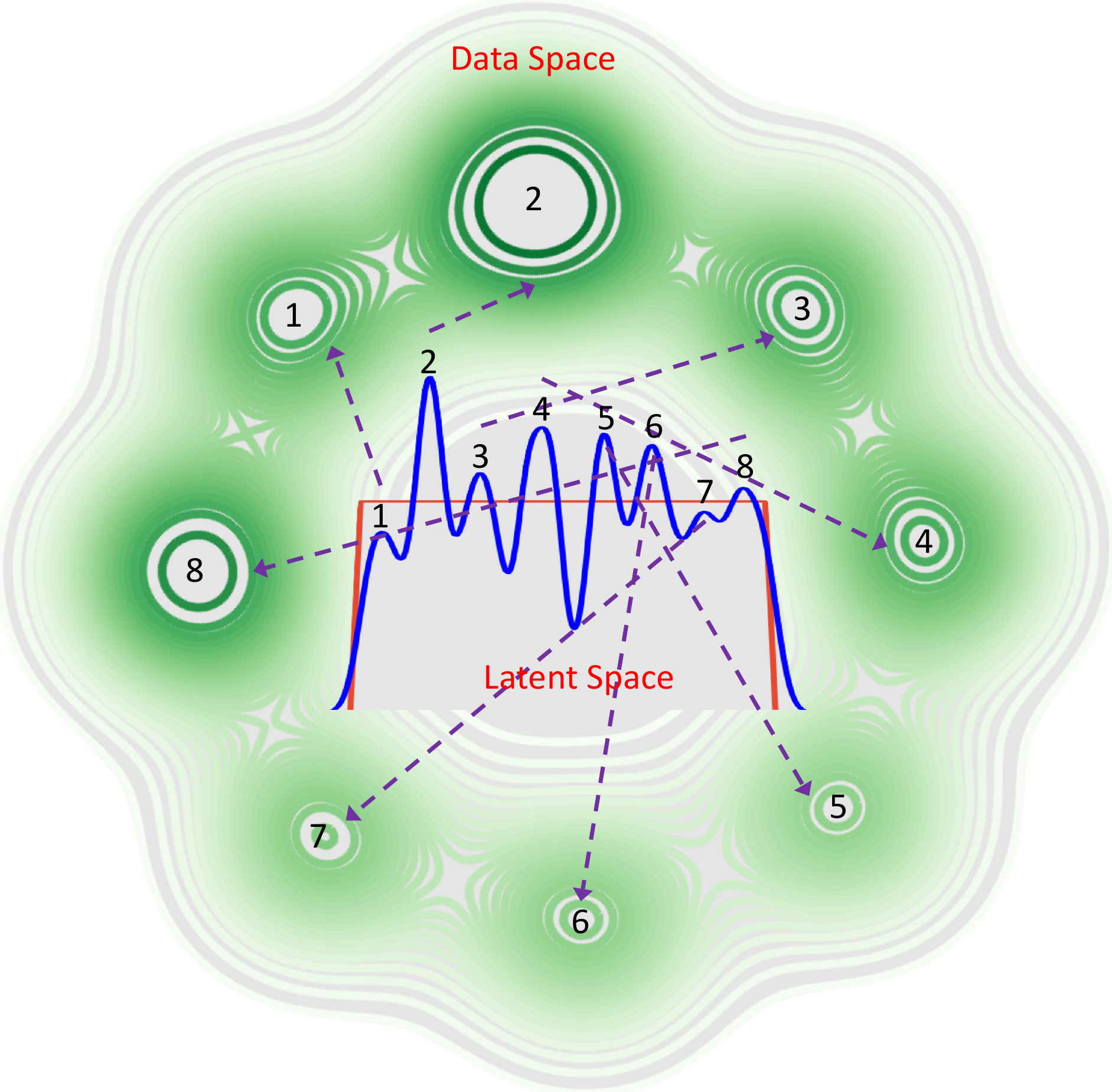}
    \caption{Latent-distribution morphing can learn an eight-mode distribution, with each mode corresponding to one mode of the target distribution.}
    \label{fig:toy_latent}
\end{figure}

Figure \ref{fig:toy_latent} shows the latent distribution after morphing, which, interestingly, also consists of 8 modes, with each corresponding to one mode in the data space. 
With such a correspondence, the mapping from the latent space to data space become much more smooth and easier to be learned. 

We also evaluate the training time of different models, which are provided in Table \ref{tab:toy_exp} in the Appendix. Although EBM trained with 30-step sampling can capture all the modes, it is much more time-consuming compared to others. When the EBM is trained with a short sampling procedure, it fails to learning all the modes of the target distribution. By contrast, our proposed model and GANs are much more efficient. 

\begin{table*}[ht!]
    \centering
    \caption{Comparisons of our model with related baselines.}
    \label{tab:generative_model_main}
        \begin{sc}
        \begin{adjustbox}{scale=0.65}
            \begin{tabular}{lcccccc}
            \toprule
            & & &\multicolumn{2}{c}{CIFAR-10 ($32\times32$)} & \multicolumn{2}{c}{STL-10 ($48\times48$)}\\
            & Morphing/Sampling & Generator & FID $(\downarrow)$ & IS $(\uparrow)$ & FID $(\downarrow)$ & IS $(\uparrow)$\\
            \midrule
            &\multicolumn{6}{c}{CNN-based}\\
            WGAN-GP  \citep{gulrajani2017improved} & &  \checkmark & $28.42 \pm 0.06$ & $7.14 \pm 0.09$ & $39.88 \pm 0.06$ & $8.85 \pm 0.09$ \\
            SN-GAN \citep{miyato2018spectral} & &  \checkmark & $22.30 \pm 0.12$ & $7.54 \pm 0.09$ & $40.45 \pm 0.17$ & $8.45 \pm 0.08$\\
            SN-SMMD-GAN \citep{arbel2018gradient} & &  \checkmark & $25.0 \pm 0.3$ & $7.3 \pm 0.1$ & $40.6 \pm 0.1$ & $8.5 \pm 0.1$\\
            CR-GAN \citep{zhang2019consistency} & &  \checkmark & $18.72$ & $7.93$& -- & -- \\
            Improved MMD-GAN \citep{wang2018improving} & &  \checkmark & $16.21$ & $8.29$ & $36.67$ & $9.36$ \\
            DOT \citep{tanaka2019discriminator} & \checkmark &  \checkmark & $15.78 $ & $8.50 \pm 0.01$ & $34.84 $ & $10.30 \pm 0.21$ \\
            DG$f$low \citep{ansari2020refining} & \checkmark &  \checkmark & $15.30 \pm 0.08 $ & $8.14 \pm 0.03$ & $34.60 \pm 0.11$ &  $10.41 \pm 0.02$  \\
            \textbf{Ours  (trained w/o morphing)} & \checkmark &  \checkmark & $9.71 \pm 0.05$ & $9.24 \pm 0.10$ & $22.15 \pm 0.12$ & $11.12 \pm 0.10$\\
            \textbf{Ours}& \checkmark &  \checkmark & $\mathbf{9.01 \pm 0.07}$ & $\mathbf{9.38 \pm 0.13}$ & $\mathbf{21.40 \pm 0.05}$ & $\mathbf{11.17 \pm 0.09}$\\
            \midrule
            &\multicolumn{6}{c}{ResNet-based}\\
            EBM \citep{NIPS2019_8619} & \checkmark &   & $40.58$ & $6.02$ & - & -\\
            NCSN \citep{song2019generative} & \checkmark &   & $25.32$ & $8.87 \pm 0.12$ & - & -\\
            SN-GAN \citep{miyato2018spectral} & &  \checkmark & $21.70 \pm 0.21$ & $8.22 \pm 0.05$ & $40.10 \pm 0.50$ & $9.10 \pm 0.04$\\
            CR-GAN \citep{zhang2019consistency} & &  \checkmark & $14.56$ & $8.40$  & -- & -- \\
            BigGAN \citep{brock2018large} & &  \checkmark & $14.73$ & $9.22$  & -- & -- \\
            Auto-GAN \citep{gong2019autogan} & &  \checkmark & $12.42$ & $8.56 \pm 0.10$ & $31.01$ & $9.16 \pm 0.12$ \\
            AdversarialNAS-GAN \citep{Gao_2020_CVPR} & &  \checkmark & $10.87$ & $8.74 \pm 0.07$ & $26.98$ & $9.63 \pm 0.19$ \\
            DOT \citep{tanaka2019discriminator} & \checkmark &  \checkmark & $19.71 $ & $8.50 \pm 0.12$ & $39.48 $ & $10.03 \pm 0.14$ \\
            DDLS \cite{che2020your} & \checkmark &  \checkmark & $15.42 $ & $9.09 \pm 0.10$ & - &  -  \\
            DG$f$low \citep{ansari2020refining} & \checkmark &  \checkmark & $9.62 \pm 0.03 $ & $\mathbf{9.35 \pm 0.03}$ & - &  -  \\
            \textbf{Ours} & \checkmark &  \checkmark & $ \mathbf{7.95 \pm 0.02 }$ & $ 9.24 \pm 0.11$ & $\mathbf{21.23 \pm 0.05 }$ & $\mathbf{11.17 \pm 0.12}$\\
            \midrule
            &\multicolumn{6}{c}{StyleGAN2-based}\\
            StyleGAN2 \citep{karras2020analyzing} & & \checkmark & $8.32 \pm 0.09$ & $9.21 \pm 0.09$ & - & - \\
            StyleGAN2+ADA \citep{karras2020training} & & \checkmark & $2.92 \pm 0.05$ & $9.83 \pm 0.04$  & - & -  \\
            \textbf{Ours} & \checkmark & \checkmark & $\mathbf{2.82 \pm 0.04}$ & $\mathbf{10.07 \pm 0.09}$ & - & -  \\
            \bottomrule
            \end{tabular}  
        \end{adjustbox}
        \end{sc}
\end{table*}
\subsection{Image Generation}
\paragraph{Main results} 
Following previous works \citep{miyato2018spectral, karras2020analyzing}, we conduct image generation experiments on three model achitectures: CNN-based, ResNet-based, and StyleGAN2-based models. The CNN-based model and ResNet-based model follow the architecture design as \cite{miyato2018spectral}, while the StyleGAN2-based model follows \citep{karras2020analyzing}.
For fair comparisons, we use the KL divergence as the functional for optimization in \eqref{eq:grad_final_kernel}, as other functionals will need an extra network as the variational function. Meanwhile, the functional used in \eqref{eq:latent_sampling} can be freely chosen from Table \ref{tab:different_update_rule}. Fr\'echet inception distance (FID) and Inception Score (IS) are reported following previous works. We compare our method with popular baselines, 
some of the baselines such as DG$f$low, StyleGAN2+ADA represent the state-of-the-arts in their settings.

The quantitative results are reported in Table \ref{tab:generative_model_main}. Results of the CNN-based model are from two settings: training without latent-distribution morphing and with a 5-step morphing. The ResNet-based model is trained without morphing, and the StyleGAN2-based model is based on the pre-trained model by \citep{karras2020training}. Our proposed method obtains the best results under different architectures, which outperforms the CNN-based generative model by a large margin, and is also better than other ResNet-based generative models including the ones with Neural Architecture Search (Auto-GAN and AdversarialNAS-GAN).

\begin{table}[t!]
    \caption{Plug-and-play on GAN-based models.}
    \centering
    \label{tab:arbitrary_gan}
        \begin{sc}
        \begin{adjustbox}{scale=0.63}
            \begin{tabular}{lcccr}
            \toprule
            &\multicolumn{2}{c}{CIFAR-10 ($32\times32$)} & \multicolumn{2}{c}{STL-10 ($48\times48$)}\\
            Models& FID $(\downarrow)$ & IS $(\uparrow)$ & FID $(\downarrow)$ & IS $(\uparrow)$\\
            \midrule
            SN-MMD-GAN & $28.49 \pm 0.17$ & $7.18 \pm 0.10$ & $49.59 \pm 0.06$ & $7.92 \pm 0.07$\\
            \quad + morphing & $23.91 \pm 0.10$ & $8.26 \pm 0.06$ & $42.05 \pm 0.11$ & $9.19 \pm 0.08$ \\
            \quad + regularizer & $14.60 \pm 0.05$ & $\mathbf{8.61 \pm 0.11}$ & $38.89 \pm 0.06$ & $9.22 \pm 0.04$\\
            \midrule 
            SN-GAN & $22.30 \pm 0.12$ & $7.54 \pm 0.09$ & $40.45 \pm 0.17$ & $8.45 \pm 0.08$\\
            \quad + morphing & $14.91 \pm 0.09$ & $8.28 \pm 0.11$ & $34.10 \pm 0.12$ & $9.13 \pm 0.10$ \\
            \quad + regularizer & $\mathbf{14.45 \pm 0.04}$ & $8.38 \pm 0.06$ & $\mathbf{33.79\pm 0.05}$ & $9.25 \pm 0.07$\\
            \midrule 
            WGAN-GP & $28.42 \pm 0.06$ & $7.14 \pm 0.09$ & $39.88 \pm 0.06$ & $8.85 \pm 0.09$ \\
            \quad + morphing & $21.78 \pm 0.13$ & $7.89 \pm 0.08$ & $35.57\pm 0.05$ & $9.33 \pm 0.09$\\
            \quad + regularizer & $19.15 \pm 0.07$ & $8.09 \pm 0.11$ & $34.24 \pm 0.12$ & $\mathbf{9.38 \pm 0.05}$\\
            \midrule 
            SN-Sinkhorn-GAN & $26.36 \pm 0.15$ & $7.08 \pm 0.08$ & $53.27 \pm 0.14$ & $7.58 \pm 0.07$ \\
            \quad + morphing & $23.08 \pm 0.07$ & $7.77 \pm 0.09$ & $49.40 \pm 0.12$ & $8.14 \pm 0.08$ \\
            \quad + regularizer &  $21.15 \pm 0.09$ & $7.83 \pm 0.11$ & $45.44 \pm 0.11$ & $8.32 \pm 0.10$\\
            \bottomrule
            \end{tabular}  
        \end{adjustbox}
        \end{sc}
\end{table}

\begin{figure*}[t!]
    \centering
    \subfigure{\includegraphics[width=0.98\linewidth]{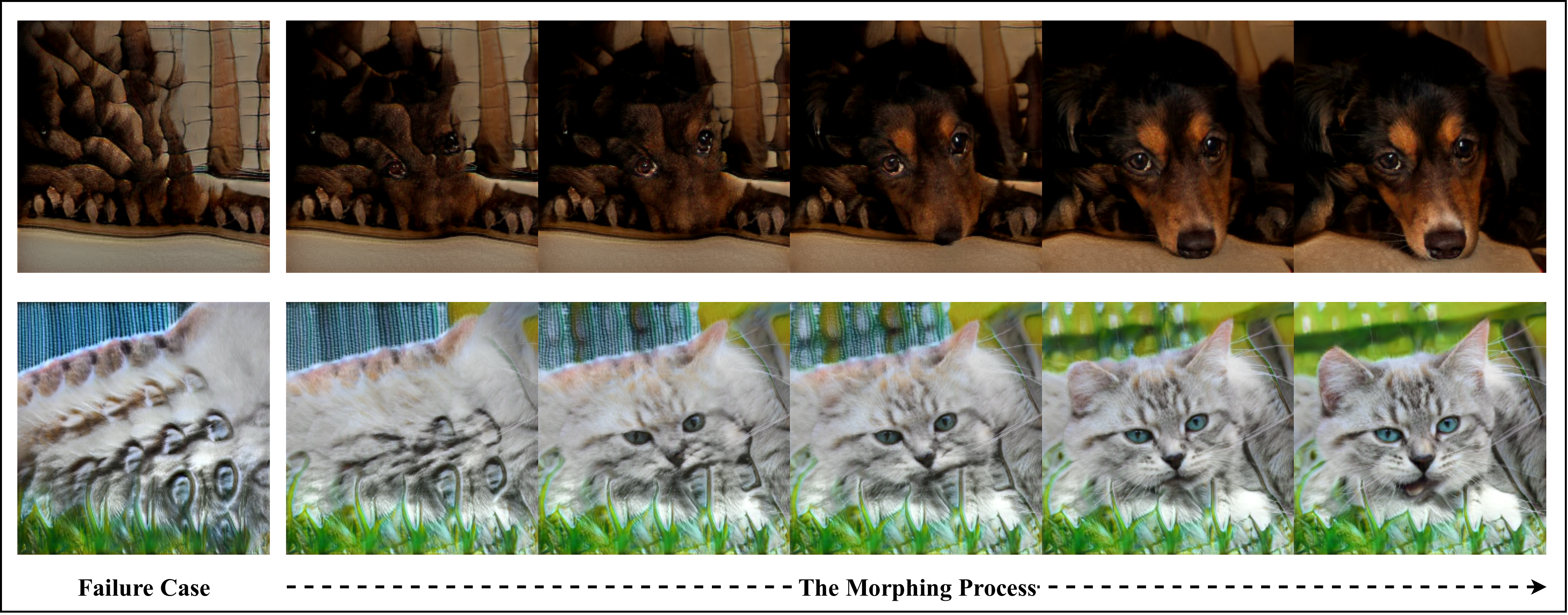}}
    \vspace{-0.3cm}
    \caption{Failure cases mitigation with our latent-distribution morphing on the state-of-the-art StyleGAN2+ADA. First column: failure case on AFHQ Dog ($512\times512$) and AFHQ Cat ($512\times512$), respectively.}
    \label{fig:improvement}
    \vspace{-0.3cm}
\end{figure*}

\paragraph{Plug-and-play on pre-trained GANs} 
We show how to improve pre-trained GANs with our latent-distribution morphing scheme for plug-and-play during testing. All the models are constructed using a 4-layer CNN generator and a 7-layer CNN discriminator as in  \cite{miyato2018spectral}. 
The discriminator output dimensions of SN-MMD-GAN and SN-Sinkhorn-GAN \citep{genevay2018learning} are set to 16, while the dimensions are set to 1 for others. 
Quantitative results are provided in Table \ref{tab:arbitrary_gan}. 
Our proposed method indeed improves the performance, which is better than related works \citep{tanaka2019discriminator, che2020your, ansari2020refining}.
Our plug-and-play scheme can also improve the current state-of-the-arts model trained on high-resolution images.
For example, although StyleGAN2+ADA \citep{karras2020training} is able to generate high-resolution images 
with impressive details, there also exist many failure cases as shown in Figure \ref{fig:improvement}. 
With our latent-distribution morphing scheme, many of these failure cases can be recovered, as shown in Figure \ref{fig:improvement}. More results are provided in the Appendix. 
Specifically, the morphing process gradually refines the details while preserving some global information such as the position, foreground and background colors, {\it etc}. 
Some quantitative results are also provided in Table \ref{tab:style_gan}.

\begin{table}[ht!]
\centering
    \caption{Extension to EBMs.}
    \label{tab:ebm_extension}
    \begin{sc}
    \hspace{-0.6cm}
    \begin{adjustbox}{scale=0.65}
    \begin{tabular}{lcc}
        \toprule
         \multicolumn{3}{c}{CIFAR-10 ($32\times32$)}\\
         Models & FID $(\downarrow)$ & IS $(\uparrow)$ \\
         \midrule
         \multicolumn{3}{c}{without generator}\\
         EBM \citep{NIPS2019_8619} & $37.90$ & $8.30$ \\
         $f$-EBM \citep{yu2020training} & $ 30.86$ & $8.61 \pm 0.06$ \\
         Ours (EBM) & $35.50 \pm 0.03$ & $8.60 \pm 0.09$ \\
         \textbf{Ours ($f$-EBM)} & $\mathbf{28.84 \pm 0.04}$ & $\mathbf{8.82 \pm 0.07}$\\
         \midrule
          \multicolumn{3}{c}{with generator}\\
            MEG \citep{kumar2019maximum}& $33.18$ & $7.31 \pm 0.06$ \\
            VERA \citep{grathwohl2020nomcmc} &$27.5$ & $8.00$\\
            GEBM \citep{Arbel2020KALEWE} & $19.32$ & --\\
          \textbf{Ours} & $\mathbf{7.95 \pm 0.02}$ & $\mathbf{9.24 \pm 0.11}$ \\
         \bottomrule
    \end{tabular}
    \end{adjustbox}
    \end{sc}
\end{table}

\paragraph{Kernel-based EBMs}
As described previously, we can construct a kernel-based variant of vanilla EBM \citep{NIPS2019_8619} or $f$-EBM  \citep{yu2020training} by removing the generator. We adopt the network architecture in \cite{NIPS2019_8619} for fairness. We can either train an EBM from scratch or directly improve the pre-trained EBM with our morphing scheme. The comparisons are provided in Table \ref{tab:ebm_extension}, we also compare our model with other methods which combine EBMs with generators.
As expected, 
our kernel-based EBM obtains better results than the standard EBM and $f$-EBM.

\begin{table}[t!]
    \centering
    \caption{FID results of the proposed method on CIFAR-10.}
    \label{tab:different_sampling_cnn_cifar10}
    \begin{sc}
    \begin{adjustbox}{scale=0.65}
    \begin{tabular}{lccccc}
        \toprule
         \multirow{2}{*}{Steps} &  \multicolumn{5}{c}{Functional}\\
         & Langevin & KL &  RKL & JS & SH \\
         \midrule 
          \multicolumn{6}{c}{Trained without morphing} \\
          0 &  \multicolumn{5}{c}{$16.78 \pm 0.08$} \\
         10 & $11.72 \pm 0.07$ & $10.45 \pm 0.06$ & $12.37 \pm 0.08$ & $11.09 \pm 0.04$ & $11.08 \pm 0.08$\\
         20 & $11.84 \pm 0.10$ & $10.12 \pm 0.10$ & $11.16 \pm 0.10$ & $10.44 \pm 0.11$ & $10.37 \pm 0.07$\\
         30 & $11.69 \pm 0.15$ & $9.81 \pm 0.04$ & $10.75 \pm 0.06$ & $9.89 \pm 0.03$ & $9.95 \pm 0.06$\\
         40 & $11.56 \pm 0.14$ & $9.88 \pm 0.09$ & $10.51 \pm 0.07$ & $9.84 \pm 0.05$ & $\mathbf{9.71 \pm 0.05}$\\
          \midrule 
         \multicolumn{6}{c}{Trained with 5-step morphing} \\
          0 &  \multicolumn{5}{c}{$16.85 \pm 0.05$} \\
         10 & $11.30 \pm 0.08$ & $9.60 \pm 0.05$ & $10.89 \pm 0.06$ & $10.45 \pm 0.04$ & $10.00 \pm 0.04$\\
         20 & $11.26 \pm 0.03$ & $9.18 \pm 0.03$ & $9.92 \pm 0.03$ & $9.56 \pm 0.06$ & $9.54 \pm 0.04$\\
         30 & $11.15 \pm 0.06$ & $9.06 \pm 0.11$ & $9.71 \pm 0.10$ & $9.28 \pm 0.07$ & $9.49 \pm 0.05$\\
         40 & $11.50 \pm 0.05$ & $\mathbf{9.01 \pm 0.07}$ & $9.63 \pm 0.03$ & $9.27 \pm 0.05$ & $9.32 \pm 0.04$\\
         \bottomrule
    \end{tabular}
    \end{adjustbox}
    \end{sc}
    \vspace{-0.3cm}
\end{table}
\paragraph{Ablation studies}
We conduct ablation studies to investigate the following questions: How do different $f$-divergence forms in latent-distribution morphing impact the model performance, and how does the number of updates in morphing during testing influence the model performance? 
The main results are provided in Table \ref{tab:different_sampling_cnn_cifar10} (also Table \ref{tab:different_sampling_cnn_stl10} in the Appendix). From \ref{tab:different_sampling_cnn_cifar10}, we can conclude that latent-distribution morphing during training can indeed improve the model performance. However, the morphing could cost a little time overhead. Fortunately, we see from Table \ref{tab:different_sampling_cnn_cifar10} that only performing latent-distribution morphing in testing can also improve the performance by a large margin. Empirically, we find that morphing according to the KL divergence seems to be less sensitive to hyper-parameters, which is our recommendation for practical use.

\subsection{Image Translation}
We apply our model for image translation based on the StarGAN v2 model \cite{9157662}. StarGAN v2 is able to perform high-resolution image translation over multiple domains, which consists of a generator, a mapping network, a style encoder, and a discriminator. 
The generator consists of an encoder and a decoder. To translate an image from a source domain to a target domain, the encoder takes the source image as input and outputs a latent feature $\zb$. Then $\zb$ and a style code $\sbb$ are fed into the decoder to generate a translated image. Specifically, the style code $\sbb$ is extracted by two means, corresponding to the so-called latent-guided synthesis and reference-guided synthesis. 

We apply the latent-distribution morphing to StarGAN v2 to update both the style code $\sbb$ and latent code $\zb$. Following \cite{9157662}, we conduct image translation experiment on CelebA-HQ and AFHQ by evaluating the Fr\'echet inception distance (FID) and learned perceptual image patch similarity (LPIPS). All images are resized to $256\times256$. The main results are provided in Table \ref{tab:image_translation_latent} and Table \ref{tab:image_translation_reference}, in which MUNIT \citep{huang2018multimodal}, DRIT \citep{lee2018diverse}, MSGAN \citep{mao2019mode}, StarGAN v2 serve as the baseline models. Again, our method achieves the best results. More results are provided in the Appendix.

\begin{table}[t!]
    \centering
    \caption{Results of latent-guided synthesis.}
    \label{tab:image_translation_latent}
    \begin{sc}
    \begin{adjustbox}{scale=0.65}
    \begin{tabular}{lcccc}
        \toprule
         &\multicolumn{2}{c}{CelebA-HQ ($256\times256$)} & \multicolumn{2}{c}{AFHQ ($256\times256$)}\\
         Models & FID $(\downarrow)$ & LPIPS $(\uparrow)$ & FID $(\downarrow)$ & LPIPS $(\uparrow)$ \\
         \midrule
         MUNIT & $31.4$ & $0.363$ & $41.5$ & $0.511$\\
         DRIT & $52.1$ & $0.178$ & $95.6$ & $0.326$ \\
         MSGAN & $33.1$ & $0.389$ & $61.4$ & $0.517$ \\
         StarGAN v2 & $13.7$ & $0.452$ & $16.2$ & $0.450$ \\
         \midrule 
         \textbf{Ours} &  $\mathbf{13.3}$ & $\mathbf{0.480}$ & $\mathbf{15.8}$ & 0.472\\
         \bottomrule
    \end{tabular}
    \end{adjustbox}
    \end{sc}
    \vspace{-0.3cm}
\end{table}

\begin{table}[t!]
    \centering
    \caption{Results of reference-guided synthesis.}
    \label{tab:image_translation_reference}
    \begin{sc}
    \begin{adjustbox}{scale=0.65}
    \begin{tabular}{lcccc}
        \toprule
         &\multicolumn{2}{c}{CelebA-HQ ($256\times256$)} & \multicolumn{2}{c}{AFHQ ($256\times256$)}\\
         Models & FID $(\downarrow)$ & LPIPS $(\uparrow)$ & FID $(\downarrow)$ & LPIPS $(\uparrow)$ \\
         \midrule
         MUNIT & $107.1$ & $0.176$ & $223.9$ & $0.199$\\
         DRIT & $53.3$ & $0.311$ & $114.8$ & $0.156$ \\
         MSGAN & $39.6$ & $0.312$ & $69.8$ & $0.375$ \\
         StarGAN v2 & $23.8$ & $0.388$ & $19.8$ & $0.432$ \\
         \midrule 
         \textbf{Ours}  &  $\mathbf{17.9}$ & $\mathbf{0.488}$ & $\mathbf{19.2}$ & $\mathbf{0.457}$\\
         \bottomrule
    \end{tabular}
    \end{adjustbox}
    \end{sc}
\end{table}

\section{Conclusion}
We propose a new framework to learn high-dimensional distributions via a parameterized latent FP kernel. 
Our model not only achieves better performance than related models, but also improves arbitrary GAN and EBM by a plug-and-play mechanism. Extensive experiments have shown the effectiveness of our proposed method on various tasks.

\clearpage
\bibliography{main}
\bibliographystyle{icml2021}


\clearpage
\onecolumn
\appendix

\section{Proof of Lemma \ref{thm:convergence}}
\textit{
Assume that $(q^t)_{t\geq 0}$ evolves according to the WGF of $f$-divergence, $F(q^t, p) = D_{f}(p \Vert q^t)$, with initialization $q^0$. Let $d_{W_2}(\cdot, \cdot)$ be the 2-Wasserstein distance. Assume that $F(q^0, p)<\infty$ and $d_{W_2}(q^0, p)<\infty$. Then, the following holds:\\
1) $F(q^t, p)$ is non-increasing and converges to the global optimum as $t\rightarrow \infty$;\\
2) If $\Vert \nabla_{W_2} F(q^t, p) \Vert ^2 \geq \lambda F(q^t, p)$ with some constant $\lambda >0$, then $F(q^t, p) \leq \exp(-\lambda t) F(q^0, p)$;\\
3) If $F(q^t, p)$ is $\lambda$-geodesically convex with some constant $\lambda>0$, then $d_{W_2}(q^t, p) \leq \exp(-\lambda t)d_{W_2}(q^0, p)$;\\
4) If mapping $g$ is Lipschitz continuous, then $d_{W_2}(Q^t, P)$ converges with the same convergence rate as $d_{W_2}(q^t, p)$.
}
\begin{proof}
\\
1) \\
\begin{align*}
    \dfrac{\partial F(q^t, p)}{\partial t} = \langle \nabla_{W_2}F(q^t, p), \dfrac{\partial q^t}{\partial t} \rangle.
\end{align*}
According to the definition of Wasserstein gradient flow, we have $\dfrac{\partial q^t}{\partial t} = -\nabla_{W_2}F(q^t, p)$. Thus, we get 
\[
 \dfrac{\partial F(q^t, p)}{\partial t}  = - \Vert\nabla_{W_2}F(q^t, p)\Vert^2 \leq 0, 
\] 
which means that $F(q^t, p)$ is non-increasing. Since $F(q^t, p)$ is $f$-divergence, we know that it will converge to a global optimum due to the convexity property, in which case $q^{\infty} = q^{*} = p, F(q^*) = 0$.

2)\\
For simplicity, we let $F(q^t, p) = \mathcal{F}(t)$. Define $v(t) = \exp(\int_0^t -\lambda s \text{d}s) = \exp(-\lambda t)$. Then, we have
\[
v^\prime(t) = \exp(-\lambda t) \cdot(-\lambda t)^\prime = -\lambda v(t).
\]
We also know that $v(t)>0$. Thus, $v^\prime(t)<0$ for all $t$. Hence, we have
\begin{align*}
    \dfrac{\text{d}}{\text{d} t} \dfrac{\mathcal{F}(t)}{v(t)} &= \dfrac{\mathcal{F}^\prime(t) v(t) - v^\prime(t) \mathcal{F}(t)}{(v(t))^2}\\
    & = \dfrac{\mathcal{F}^\prime (t) v(t) +\lambda v(t)\mathcal{F}(t)}{(v(t))^2}\\
    &\leq \dfrac{-\lambda \mathcal{F}(t) v(t) +\lambda v(t)\mathcal{F}(t)}{(v(t))^2}\\
    &= 0,
\end{align*}
which means that $\dfrac{\mathcal{F}(t)}{v(t)}$ is non-increasing and bounded by $\dfrac{\mathcal{F}(0)}{v(0)} = \mathcal{F}(0)$. Thus, we have
\[
F(q^t, p) = \mathcal{F}(t) \leq v(t) \mathcal{F}(0) =  \exp(-\lambda t)F(q^0, p).
\]

3)
\begin{definition}[$\lambda$-geodesically convex]
Let $d(\cdot, \cdot)$ be the distance. 
A functional $\mathcal{F}$ is $\lambda$-geodesically convex if for any $v_0, v_1 \in \text{Domain}(\mathcal{F})$, there exists a geodesics $\gamma$ with $\gamma_0=v_0, \gamma_1=v_1, d(\gamma_s, \gamma_t) = d(\gamma_0, \gamma_1)(t-s), \forall 0 \leq t \leq s \leq 1$ such that
\[
\mathcal{F}(\gamma_t) \leq (1-t)\mathcal{F}(\gamma_0) + t\mathcal(F)(\gamma_1) - \dfrac{1}{2}\lambda t(1-t)d^2(\gamma_0, \gamma_1).
\]
\end{definition}

If $F(q^t, p)$ is a $\lambda$-geodesically convex functional, we can directly apply Theorem 11.1.4 in \cite{ambrosio2008gradient}, by setting one of the probability measure in Theorem 11.1.4 to be the measure associated with $p$, i.e. global optimum.

4)\\
We have 
\[
d_{W_2}(P, Q) \leq \sqrt{\int \Vert \xb - \xb^\prime \Vert^2 \text{d}\Pi} = \sqrt{\int \Vert g(\zb) - g(\zb^\prime) \Vert^2 \text{d}\pi} \leq \sqrt{\int L^2\Vert \zb - \zb^\prime \Vert^2 \text{d}\pi} = L d_{W_2}(p, q),
\]
where $\pi$ denotes the optimal transportation plan on $\mathcal{Z}^2$ with marginal $p, q$, $\Pi$ denotes the corresponding plan in $\mathcal{X}^2$, $L$ is a constant. $\Pi$ may not be the optimal plan in $\mathcal{X}^2$, thus we have the first inequality. By the Lipchistz continuity, we have the second inequality. The last equality comes from the definition of Wasserstein distance.
Now we can easily conclude the convergence rate.
\end{proof}

\section{Details on Table \ref{tab:different_update_rule}}

We would like to introduce some examples on the first variation of some functionals \citep{santambrogio2016}, which will be used to help us get Table \ref{tab:different_update_rule}.

Assume that $f: R \rightarrow R$, $V: \Omega \rightarrow R$ and $W: R^d \rightarrow R$ are regular enough, while $W$ is symmetric. Then, the first variation of the following functionals:
\[
\mathcal{F}(\rho) = \int (f(\rho))) \text{d}\xb, \mathcal{V}(\rho) = \int V(\xb) \text{d}\rho, \mathcal{W}(\rho) = \dfrac{1}{2}\int \int W(\xb - \yb) \text{d}\rho(\xb) \text{d} \rho(\yb)
\]
are
\[
\dfrac{\delta \mathcal{F}}{\delta \rho}(\rho) = f^\prime (\rho), \dfrac{\delta \mathcal{V}}{\delta \rho}(\rho) = V, \dfrac{\delta \mathcal{W}}{\delta \rho}(\rho) = W * \rho.
\]

Recall that in the morphing, we are trying to match the latent distribution $r(\zb)$ with target parametric distribution $q(\zb)$ (not $p(\zb)$), and we approximate $r(\zb)$ by $\sum_{i=1}^n K(g(\zb), g(\zb_i))$, and $q(\zb) \propto \int K(g(\zb), \xb^\prime) P(\xb^\prime) \text{d} (\xb^\prime) \approx \sum_{i=1}^n K(g(\zb), \xb^\prime_i)/n$.

1) In the case of KL-divergence, $F(r) = \int r \log(r / q) \text{d} \xb$. We have
\[
\dfrac{\delta F(r)}{\delta r} = \log(r/q).
\]
Thus,
\[
\nabla _{\zb} \dfrac{\delta F(r)}{\delta r} = \nabla_{\zb} \log \sum_{i=1}^n K(g(\zb), g(\zb_i)) - \nabla_{\zb}  \log \sum_{i=1}^n K(g(\zb), \xb^\prime_i).
\]

2) In the case of reverse KL-divergence,  $F(r) = \int q \log(q / r ) \text{d} \xb$. We have

\[
\dfrac{\delta F(r)}{\delta r} = - \dfrac{q}{r}.
\]
Thus,
\[
\nabla _{\zb} \dfrac{\delta F(r)}{\delta r} \propto - \nabla_{\zb} \left[  \sum_i^n K(g(\zb), \xb^\prime_i) / \sum_i^n K(g(\zb), g(\zb_i))  \right].
\]
We use "$\propto$" instead of "$=$" due to the existence of normalization constant of $q(\zb)$ and $r(\zb)$. This will not influence our method because we will choose a hyper-parameter: step size $\lambda$, which will scale the gradient and cancel the impact of the unknown constant.

3) In the case of JS-divergence,  $F(r) = \dfrac{1}{2}\int r \log\dfrac{2 r}{r + q} +  q \log\dfrac{2  q}{r + q} \text{d} \xb$. We have
\[
\dfrac{\delta F(r)}{\delta r} = \dfrac{1}{2} \log \dfrac{2 r}{r + q} - \dfrac{1}{4}.
\]
Thus, 
\[
\nabla _{\zb} \dfrac{\delta F(r)}{\delta r} = -\dfrac{1}{2}\Big \{\nabla_{\xb}\log \left[ \sum_i^n K(g(\zb), g(\zb_i)) + \sum_i^n \alpha K(g(\zb), \xb^\prime_i) \right] - \nabla_{\zb} \log 2\sum_i^n K(g(\zb), g(\zb_i))\Big \}.
\]
where $\alpha$ is a hyper-parameter because of the existence of normalization constant of $q(\zb)$ and $r(\zb)$. In practice, $\alpha = 1$ works well. Thus, we simply set $\alpha=1$ in experiments.

4) In the case of SH-divergence,  $F(r) = \int (\sqrt{r} - \sqrt{ q} )^2\text{d} \xb$. We have
\[
\dfrac{\delta F(r)}{\delta r} = 1 - \sqrt{\dfrac{q}{r}}.
\]
Thus,
\[
\nabla _{\zb} \dfrac{\delta F(r)}{\delta r} \propto - \nabla_{\zb} \sqrt{  \sum_i^n K(g(\zb), \xb^\prime_i) / \sum_i^n K(g(\zb), g(\zb_i)) }.
\]
The normalization constants of $q(\zb)$ and $r(\zb)$ are hidden in the proportion.

\section{Discussion on Training Generator}

As mentioned earlier, we train the generator $g_{\thetab}$ by minimizing $\text{MMD}_{\phib}(P, R)$ because we do not have closed form of $R(\xb)$. The gradient of MMD w.r.t $\thetab$ is
\begin{align*}
    \nabla_{\thetab} \text{MMD}_{\phib}(P, R) = & \nabla_{\thetab} \{\mathbb{E}_{R(\xb), R(\xb^\prime)} \left[K_{\phib}(\xb, \xb^\prime) \right] - 2 \mathbb{E}_{R(\xb), P(\xb^\prime)} \left[K_{\phib}(\xb,\xb^\prime) \right] + \mathbb{E}_{P(\xb), P(\xb^\prime)} \left[K_{\phib}(\xb,\xb^\prime) \right] \} \\
    = & \nabla_{\thetab} \{\mathbb{E}_{r(\zb), r(\zb^\prime)} \left[K_{\phib}(g_{\thetab}(\zb), g_{\thetab}(\zb^\prime)) \right] - 2 \mathbb{E}_{r(\zb), P(\xb^\prime)} \left[K_{\phib}(g_{\thetab}(\zb),\xb^\prime) \right] + \mathbb{E}_{P(\xb), P(\xb^\prime)} \left[K_{\phib}(\xb,\xb^\prime) \right]\} \\
    = &  \mathbb{E}_{r(\zb), r(\zb^\prime)} \left[\nabla_{\thetab} K_{\phib}(g_{\thetab}(\zb), g_{\thetab}(\zb^\prime)) \right] - 2 \mathbb{E}_{r(\zb), P(\xb^\prime)} \left[\nabla_{\thetab} K_{\phib}(g_{\thetab}(\zb),\xb^\prime) \right]. 
\end{align*}
We assume that $R(\xb) = R(g_{\thetab}(\zb)) \propto r(\zb) = \mathbb{E}_{r(\zb^\prime)}\left[ K_{\phib} (g_{\thetab}(\zb), g_{\thetab}(\zb^\prime))\right]$. We rewrite this as
\[
R(g_{\thetab}(\zb)) = \dfrac{\mathbb{E}_{r(\zb^\prime)}\left[ K_{\phib} (g_{\thetab}(\zb), g_{\thetab}(\zb^\prime))\right]}{C},
\]
where $C = \int \mathbb{E}_{r(\zb^\prime)}\left[ K_{\phib} (g_{\thetab}(\zb), g_{\thetab}(\zb^\prime))\right] \text{d}\zb$ is the normalization constant. The gradient of negative log-likelihood w.r.t $\thetab$ is
\begin{align*}
    &-\nabla_{\thetab} \mathbb{E}_{P(\xb)} \left[ \log R(\xb) \right] \\
    = & -\nabla_{\thetab} \mathbb{E}_{p(\zb)} \left[ \log R(g_{\thetab}(\zb) \right] \\
    = & -\mathbb{E}_{p(\zb)} \{\nabla_{\thetab}  \log \mathbb{E}_{r(\zb^\prime)}\left[ K_{\phib} (g_{\thetab}(\zb), g_{\thetab}(\zb^\prime))\right] - \nabla_{\thetab} \log C\}\\
    = & -\mathbb{E}_{P(\xb)} \{ \nabla_{\thetab}  \log \mathbb{E}_{r(\zb^\prime)}\left[ K_{\phib} (\xb, g_{\thetab}(\zb^\prime))\right]\}  + \mathbb{E}_{P(\xb)}\left[\nabla_{\thetab} \log C\right] \\
    = & -\mathbb{E}_{P(\xb)} \Big \{ \dfrac{\nabla_{\thetab} \mathbb{E}_{r(\zb^\prime)}\left[ K_{\phib} ( g_{\thetab}(\zb^\prime), \xb)\right]}{\mathbb{E}_{r(\zb^\prime)}\left[ K_{\phib} (\xb, g_{\thetab}(\zb^\prime))\right]}\Big \}  + \mathbb{E}_{P(\xb)}\left[\dfrac{\nabla_{\thetab} C}{C}\right] \\
    = & -\mathbb{E}_{P(\xb)} \Big \{ \dfrac{\mathbb{E}_{r(\zb^\prime)}\left[ \nabla_{\thetab} K_{\phib} ( g_{\thetab}(\zb^\prime), \xb)\right]}{\mathbb{E}_{r(\zb^\prime)}\left[ K_{\phib} (\xb, g_{\thetab}(\zb^\prime))\right]}\Big \}  + \mathbb{E}_{P(\xb)}\left[\dfrac{\int \nabla_{\thetab}\int r(\zb^\prime)K_{\phib} (g_{\thetab}(\zb), g_{\thetab}(\zb^\prime)) \text{d}\zb^\prime \text{d}\zb }{C}\right] \\
    = & -\mathbb{E}_{P(\xb)} \Big \{ \dfrac{\mathbb{E}_{r(\zb^\prime)}\left[ \nabla_{\thetab} K_{\phib} ( g_{\thetab}(\zb^\prime), \xb)\right]}{\mathbb{E}_{r(\zb^\prime)}\left[ K_{\phib} (g_{\thetab}(\zb^\prime), \xb)\right]}\Big \}  + \mathbb{E}_{P(\xb)}\left[\dfrac{\int \int r(\zb^\prime) K_{\phib} (g_{\thetab}(\zb), g_{\thetab}(\zb^\prime)) \text{d}\zb^\prime \nabla_{\thetab}\log(\int r(\zb^\prime) K_{\phib} (g_{\thetab}(\zb), g_{\thetab}(\zb^\prime)) \text{d}\zb^\prime \text{d}\zb )}{C}\right] \\
    = & -\mathbb{E}_{P(\xb)} \Big \{ \dfrac{\mathbb{E}_{r(\zb^\prime)}\left[ \nabla_{\thetab} K_{\phib} ( g_{\thetab}(\zb^\prime), \xb)\right]}{\mathbb{E}_{r(\zb^\prime)}\left[ K_{\phib} (g_{\thetab}(\zb^\prime), \xb)\right]}\Big \}  +\int R(g_{\thetab}(\zb)) \nabla_{\thetab}\log(\int r(\zb^\prime) K_{\phib} (g_{\thetab}(\zb), g_{\thetab}(\zb^\prime)) \text{d}\zb^\prime) \text{d}\zb \\
    = & -\mathbb{E}_{P(\xb)} \Big \{ \dfrac{\mathbb{E}_{r(\zb^\prime)}\left[ \nabla_{\thetab} K_{\phib} ( g_{\thetab}(\zb^\prime), \xb)\right]}{\mathbb{E}_{r(\zb^\prime)}\left[ K_{\phib} (g_{\thetab}(\zb^\prime), \xb)\right]}\Big \}  + \mathbb{E}_{r(\zb)} \Big\{\dfrac{\mathbb{E}_{r(\zb^\prime)}\left[ \nabla_{\thetab} K_{\phib} (g_{\thetab}(\zb), g_{\thetab}(\zb^\prime))\right]}{\mathbb{E}_{r(\zb^\prime)}\left[ K_{\phib} (g_{\thetab}(\zb), g_{\thetab}(\zb^\prime))\right]}\Big\}.\\
\end{align*}
We can see that these two gradients are very similar to each other, except that the second one has extra terms scaling the gradient. Applying gradient descent with either is equivalent  to minimizing $K_{\phib}(g_{\thetab}(\zb), g_{\thetab}(\zb^\prime))$ and maximizing $K_{\phib}(g_{\thetab}(\zb), \xb)$, where $\zb, \zb^\prime \sim r(\zb), \xb \sim P(\xb)$. Intuitively, this is training the generator to generate samples that are similar to real samples (by maximizing the similarity between generated samples and  real ones), while the generated samples should be diverse enough (by minimizing the similarity between generated samples).

\section{Proof of Theorem \ref{thm:grad_of_f_divergence}}
\textit{
For distribution $q(\zb)$ parameterized as \eqref{eq:density_parameterization}, the gradient of arbitrary $f$-divergence 
$
    D_{f}(p \Vert q) = \mathbb{E}_{q(\zb)}\left[ f(p(\zb)/q(\zb))\right]
$
w.r.t. $\phib$ can be written as:
\[
    \nabla_{\phib}D_{f}(p \Vert q) = -\mathbb{E}_{q(\zb)}\Big \{f^*(v^*(\zb)) \nabla_{\phib}\log\{ \mathbb{E}_{P(\xb^\prime)}\left[ K_{\phib}(g_{\thetab}(\zb), \xb^\prime)\right]\} \Big\} + \mathbb{E}_{q(\zb)}\left[ f^*(v^*(\zb))\right] \mathbb{E}_{q(\zb)}\big \{\nabla_{\phib} \log\{\mathbb{E}_{P(\xb^\prime)} \left[ K_{\phib}(g_{\thetab}(\zb), \xb^\prime)\right]\}\big \}. 
\]
Specifically, the gradient of KL divergence $D_{KL}(p \Vert q)$ w.r.t. $\phib$ can be written as:
\[    
\nabla_{\phib}D_{KL}(p \Vert q) =  - \mathbb{E}_{P(\xb)}\{\nabla_{\phib} \log\{\mathbb{E}_{P(\xb^\prime)} \left[ K_{\phib}(\xb, \xb^\prime)\right]\}\} + \mathbb{E}_{q(\zb)}\{\nabla_{\phib} \log\{\mathbb{E}_{P(\xb^\prime)} \left[ K_{\phib}(g_{\thetab}(\zb), \xb^\prime)\right]\}\}.
\]
}
\begin{proof}
Recall that our parameterization
\[
q(\zb) = \tilde{q}(\xb) /Z=  \int K_{\phib}(g_{\thetab}(\zb), \xb) P(\xb^\prime) \text{d} \xb^\prime /Z_{\phib}, Z_{\phib}= \int \int K_{\phib}(g_{\thetab}(\zb), \xb) P(\xb^\prime) \text{d} \xb^\prime \text{d}  \zb
\]

and the variational representation of $f$-divergence \cite{nguyen2010estimating} is:
\[
D_{f}(p \Vert q) = \sup _{v \in \mathcal{V}} \mathbb{E}_{p(\zb)}\left[ v(\zb) \right] - \mathbb{E}_{q(\zb)}\left[ f^*(v(\zb))\right],
\]
where $f^*$ is the Fenchel conjugate function of $f$, and sub-differential $\partial f(p/q)$ contains an element of $\mathcal{V}$. Let $v^* \in \mathcal{V}$ be the function when supremum is attained, we have
\begin{align*}
    \nabla_{\phib} D_{f}(p \Vert q)  
    & = - \nabla_{\phib} \mathbb{E}_{q(\zb)}\left[ f^*(v^*(\zb))\right] \\
    & =  - \int \nabla_{\phib} q(\zb) f^*(v^*(\zb)) \text{d} \zb \\
    & =  - \int  q(\zb) \{\nabla_{\phib}  \log \int K_{\phib}(g_{\thetab}(\zb), \xb^\prime) P(\xb^\prime) \text{d} \xb^\prime - \nabla_{\phib}  \log Z_{\phib}\} f^*(v^*(\zb)) \text{d} \zb \\
    & =  - \int  q(\zb) f^*(v^*(\xb)) \nabla_{\phib}  \log\int K_{\phib}(g_{\thetab}(\zb), \xb^\prime) P(\xb^\prime) \text{d} \xb^\prime \text{d} \zb + \int  q(\zb)  f^*(v^*(\xb))  \nabla_{\phib}  \log Z_{\phib} \text{d} \zb  \\
    & = -\mathbb{E}_{q(\zb)}\{f^*(v^*(\zb)) \nabla_{\phib}\log\{ \mathbb{E}_{P(\xb^\prime)}\left[ K_{\phib}(g_{\thetab}(\zb), \xb^\prime)\right]\} \} + \int  q(\zb)  f^*(v^*(\zb)) \dfrac{\nabla_{\phib} Z_{\phib}}{Z_{\phib}}  \text{d} \zb  \\
    & = -\mathbb{E}_{q(\zb)}\{f^*(v^*(\zb)) \nabla_{\phib}\log\{ \mathbb{E}_{P(\xb^\prime)}\left[ K_{\phib}(g_{\thetab}(\zb), \xb^\prime)\right]\} \} + \int  q(\zb)  f^*(v^*(\zb))  \dfrac{\nabla_{\phib} \int \int K_{\phib}(g_{\thetab}(\zb^\prime), \xb^\prime) P(\xb^\prime) \text{d} \xb^\prime \text{d} \zb^\prime }{Z_{\phib}} \text{d} \zb  \\
    & =  -\mathbb{E}_{q(\zb)}\{f^*(v^*(\zb)) \nabla_{\phib}\log\{ \mathbb{E}_{P(\xb^\prime)}\left[ K_{\phib}(g_{\thetab}(\zb), \xb^\prime)\right]\} \} \\
    & \quad + \int  q(\zb)  f^*(v^*(\zb)) \text{d} \zb \cdot \dfrac{\int \tilde{q}(\zb) \nabla_{\phib} \log \int K_{\phib}(g_{\thetab}(\zb), \xb^\prime) P(\xb^\prime) \text{d} \xb^\prime \text{d} \zb}{Z_{\phib}}\\
    & = -\mathbb{E}_{q(\zb)}\Big \{f^*(v^*(\zb)) \nabla_{\phib}\log\{ \mathbb{E}_{P(\xb^\prime)}\left[ K_{\phib}(g_{\thetab}(\zb), \xb^\prime)\right]\} \Big\} + \mathbb{E}_{q(\zb)}\left[ f^*(v^*(\zb))\right] \mathbb{E}_{q(\zb)}\big \{\nabla_{\phib} \log\{\mathbb{E}_{P(\xb^\prime)} \left[ K_{\phib}(g_{\thetab}(\zb), \xb^\prime)\right]\}\big \}. 
\end{align*}
For KL-divergence, we have:
\begin{align*}
    & \nabla_{\phib}D_{KL}(p \Vert q) \\
    = &  - \nabla_{\phib} \mathbb{E}_{p(\zb)} \left[ \log  q(\zb) \right] \\
    = & - \mathbb{E}_{p(\zb)} \left[ \nabla_{\phib} \log \int K_{\phib}(g_{\thetab}(\zb), \xb^\prime) P(\xb^\prime) \text{d} \xb^\prime - \nabla_{\phib} \log Z_{\phib} \right] \\
    = & - \mathbb{E}_{p(\zb)} \left[ \nabla_{\phib} \log \int K_{\phib}(g_{\thetab}(\zb), \xb^\prime) P(\xb^\prime) \text{d} \xb^\prime\right] + \mathbb{E}_{p(\zb)} \left[ \nabla_{\phib} \log Z_{\phib} \right] \\
    = & - \mathbb{E}_{P(\xb)} \left[ \nabla_{\phib} \log \int K_{\phib}(\xb, \xb^\prime) P(\xb^\prime) \text{d} \xb^\prime\right] +  \mathbb{E}_{p(\zb)} \left[ \dfrac{\nabla_{\phib} Z_{\phib}}{Z_{\phib}} \right] \\
    = & - \mathbb{E}_{P(\xb)}\{\nabla_{\phib} \log\{\mathbb{E}_{P(\xb^\prime)} \left[ K_{\phib}(\xb, \xb^\prime)\right]\}\} + \mathbb{E}_{p(\zb)} \left[ \dfrac{\nabla_{\phib} \int \int K_{\phib}(g_{\thetab}(\zb), \xb^\prime) P(\xb^\prime) \text{d} \xb^\prime \text{d} \zb}{Z_{\phib}} \right] \\
    = & - \mathbb{E}_{P(\xb)}\{\nabla_{\phib} \log\{\mathbb{E}_{P(\xb^\prime)} \left[ K_{\phib}(\xb, \xb^\prime)\right]\}\} + \mathbb{E}_{p(\zb)} \left[ \dfrac{\int \int K_{\phib}(g_{\thetab}(\zb), \xb^\prime) P(\xb^\prime) \text{d} \xb^\prime \cdot  \nabla_{\phib} \log \int K_{\phib}(g_{\thetab}(\zb), \xb^\prime) P(\xb^\prime) \text{d} \xb^\prime \text{d} \zb}{Z_{\phib}} \right] \\
    = & - \mathbb{E}_{P(\xb)}\{\nabla_{\phib} \log\{\mathbb{E}_{P(\xb^\prime)} \left[ K_{\phib}(\xb, \xb^\prime)\right]\}\} + \mathbb{E}_{p(\zb)} \left[ \dfrac{\int \tilde{q}(\zb) \nabla_{\phib} \log \int K_{\phib}(g_{\thetab}(\zb), \xb^\prime) P(\xb^\prime) \text{d} \xb^\prime \text{d} \xb}{Z_{\phib}} \right] \\
    = & - \mathbb{E}_{P(\xb)}\{\nabla_{\phib} \log\{\mathbb{E}_{P(\xb^\prime)} \left[ K_{\phib}(\xb, \xb^\prime)\right]\}\} + \mathbb{E}_{p(\zb)} \{ \mathbb{E}_{q(\zb)}\{\nabla_{\phib} \log\{\mathbb{E}_{P(\xb^\prime)} \left[ K_{\phib}(g_{\thetab}(\zb), \xb^\prime)\right]\}\}\} \\
    = & - \mathbb{E}_{P(\xb)}\{\nabla_{\phib} \log\{\mathbb{E}_{P(\xb^\prime)} \left[ K_{\phib}(\xb, \xb^\prime)\right]\}\} + \mathbb{E}_{q(\zb)}\{\nabla_{\phib} \log\{\mathbb{E}_{P(\xb^\prime)} \left[ K_{\phib}(g_{\thetab}(\zb), \xb^\prime)\right]\}\}.
\end{align*}

\end{proof}

\section{Discussion on $f^*$ and $v^*$}
We provide some conjugate functions $f^*$ of different $f$-divergence in Table \ref{tab:conjugate_example}, readers interested in using $f$-divergence in GANs and EBMs may also refer to \cite{nowozin2016f, yu2020training}.

\begin{table}[h!]
    \centering
    \caption{Examples of $f^*$ for different $f$-divergence}
    \label{tab:conjugate_example}
    \begin{sc}
    \begin{adjustbox}{scale=0.8}
    \begin{tabular}{lccc}
        \toprule
        $f$-divergence & $D_f(p \Vert q)$ & $f(u)$ & $f^*(t)$ \\
        \midrule
         Kullback-Leibler & $\int p(\xb) \log \dfrac{p(\xb)}{q(\xb)} d\xb$ & $u \log u$ & $\exp(t-1)$ \\
         Reverse Kullback-Leibler & $\int q(\xb) \log \dfrac{q(\xb)}{p(\xb)} d\xb$ & $-\log u$ & $- 1 - \log(-t)$ \\
         Jensen-Shannon & $\dfrac{1}{2} \int p(\xb) \log \dfrac{2p(\xb)}{p(\xb) + q(\xb)} +  q(\xb) \log \dfrac{2q(\xb)}{p(\xb) + q(\xb)} d\xb$ & $-(u+1)\log \dfrac{1+u}{2} + u\log u $ & $-\log(2- \exp(t))$ \\
         Squared Hellinger & $\int (\sqrt{p(\xb)} - \sqrt{q(\xb)})^2  d\xb$ & $(\sqrt{u} - 1)^2$ & $\dfrac{t}{1-t}$\\
         \bottomrule
    \end{tabular}
    \end{adjustbox}
    \end{sc}
\end{table}

In practice, we may use a neural network to construct $v_{\omegab}$, then train the distribution parameter $\phib$ and variational parameter $\omegab$ in a min-max way:
\[
\min_{\phib} \max_{\omegab} \mathbb{E}_{p(\zb)} \left[ v_{\omegab}(\zb) \right] - \mathbb{E}_{q_{\phib}(\zb)}\left[f^*(v_{\omegab}(\zb)) \right]
\]
where the $\max_{\omegab}$ tries to obtain the supremum as in the original vairational form, $\min_{\phib}$ tries to minimize the $f$-divergence between $p$ and $q_{\phib}$. Note that there are some other variational forms of $f$-divergence \citep{yu2020training}, which can also be used under our proposed framework.

\section{Toy Experiments}
\begin{figure*}[ht!]
    \centering
    \subfigure[Results of the standard GAN.]{\includegraphics[width = 0.7\linewidth]{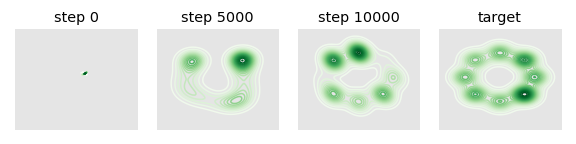}}
    \subfigure[Results of the standard GAN with the proposed morphing during testing.]{\includegraphics[width = 0.7\linewidth]{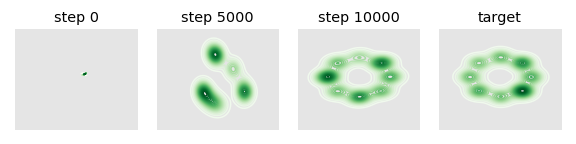}}
    \caption{More results of standard GAN in toy experiment.}
    \label{fig:toy_more_results_gan}
\end{figure*}

\begin{figure*}[ht!]
    \centering
    \subfigure[Results of the MMD-GAN.]{\includegraphics[width = 0.7\linewidth]{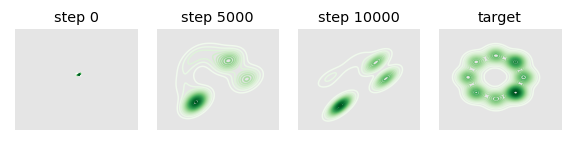}}
    \subfigure[Results of the MMD-GAN with the proposed morphing during testing.]{\includegraphics[width = 0.7\linewidth]{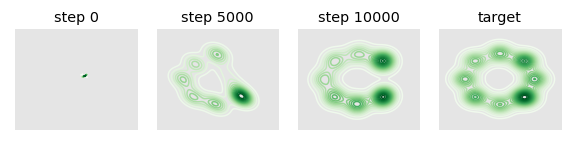}}
    \caption{More results of MMD-GAN in toy experiment.}
    \label{fig:toy_more_results_mmd_gan}
\end{figure*}

\begin{figure*}[ht!]
    \centering
    \subfigure[Results of the EBM with 10-step sampling.]{\includegraphics[width = 0.7\linewidth]{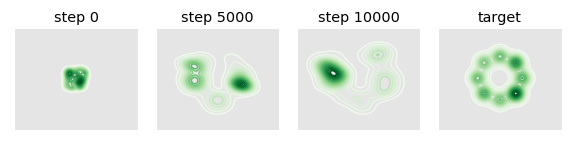}}
    \subfigure[Results of the EBM with 30-step sampling.]{\includegraphics[width = 0.7\linewidth]{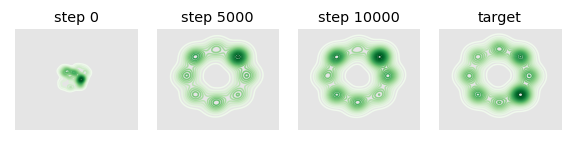}}
    \caption{More results of EBM in toy experiment.}
    \label{fig:toy_more_results_ebm}
\end{figure*}

\begin{figure*}[ht!]
    \centering
    \subfigure[Results of the proposed method.]{\includegraphics[width = 0.7\linewidth]{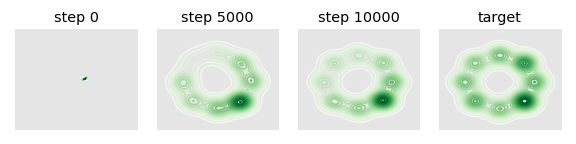}}
    \caption{More results of the proposed model in toy experiment.}
    \label{fig:toy_more_results_ours}
\end{figure*}

We present some more results of the toy experiment here. Table \ref{tab:toy_exp} shows the training time of different models in this experiment. Figure \ref{fig:toy_more_results_gan}, \ref{fig:toy_more_results_mmd_gan}, \ref{fig:toy_more_results_ebm}, \ref{fig:toy_more_results_ours} are results of different models. Our proposed model is efficient as it can learn the target distribution with little training time. With the proposed morphing method, GANs can also capture all the modes of the target distribution. Their training time slightly improves because of the use of regularizer $\mathcal{L}_{kde}$.

\begin{table}[ht!]
    \centering
    \caption{Training time of different models in toy experiment.}
    \begin{tabular}{lc}
        \toprule
         Models & Time  \\
         \midrule
         GAN & \textbf{42} s\\
         GAN + proposed morphing & 54 s\\
         \midrule
         MMD-GAN & 58 s \\
         MMD-GAN + proposed morphing & 1 min 5 s \\
         \midrule
         EBM (10-step) & 1 min 28 s \\
         EBM (30-step) & 3 min 41 s \\
         \midrule
         Ours & 59 s\\
         \bottomrule
    \end{tabular}
    \label{tab:toy_exp}
\end{table}

\clearpage

\section{More Results on Refining AFHQ Generation}

\begin{table}[h!]
    \caption{Plug-and-play StyleGAN2+ADA on AFHQ ($512\times512$)}
    \centering
    \label{tab:style_gan}
        \begin{sc}
        \begin{adjustbox}{scale=0.7}
            \begin{tabular}{lccc}
            \toprule
            & Cat & Dog & Wild \\
            Models & FID $(\downarrow)$& FID $(\downarrow)$ & FID $(\downarrow)$\\
            \midrule
            StyleGAN2+ADA & $3.55$ & $7.40$ & $3.05$\\
            \textbf{Ours} & $\mathbf{3.17}$ & $\mathbf{7.31}$ & $\mathbf{2.99}$ \\
            \bottomrule
            \end{tabular}  
        \end{adjustbox}
        \end{sc}
\end{table}

\begin{figure*}[h!]
    \centering
    \subfigure[Random seed 301]{\includegraphics[width=0.16\linewidth]{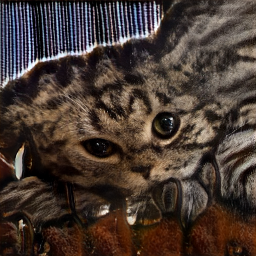}
    \includegraphics[width=0.16\linewidth]{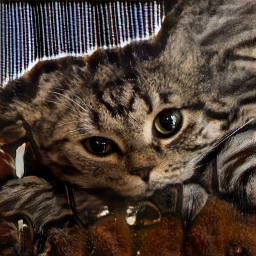}
    \includegraphics[width=0.16\linewidth]{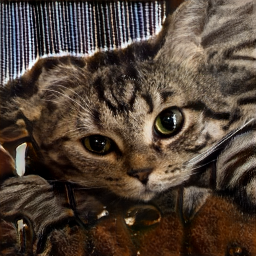}
    \includegraphics[width=0.16\linewidth]{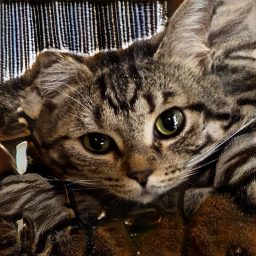}
    \includegraphics[width=0.16\linewidth]{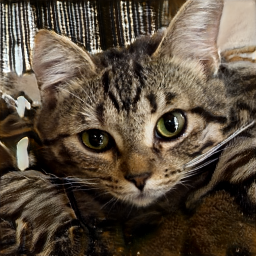}
    \includegraphics[width=0.16\linewidth]{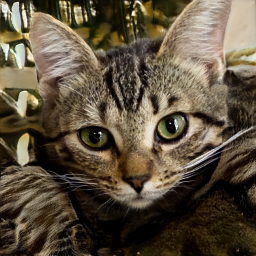}
    }
    \subfigure[Random seed 869]{\includegraphics[width=0.16\linewidth]{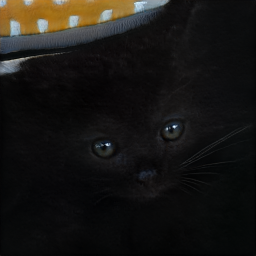}
    \includegraphics[width=0.16\linewidth]{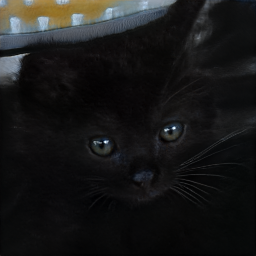}
    \includegraphics[width=0.16\linewidth]{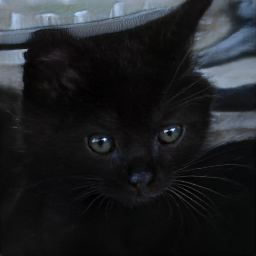}
    \includegraphics[width=0.16\linewidth]{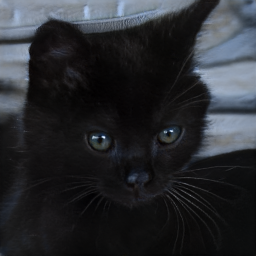}
    \includegraphics[width=0.16\linewidth]{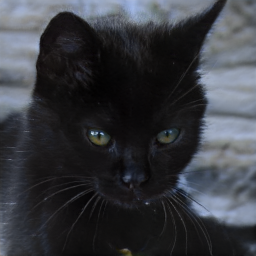}
    \includegraphics[width=0.16\linewidth]{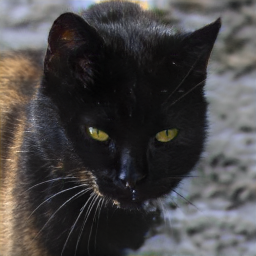}
    }
    \subfigure[Random seed 1348]{\includegraphics[width=0.16\linewidth]{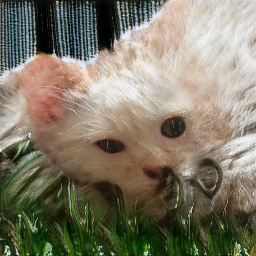}
    \includegraphics[width=0.16\linewidth]{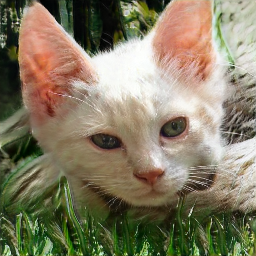}
    \includegraphics[width=0.16\linewidth]{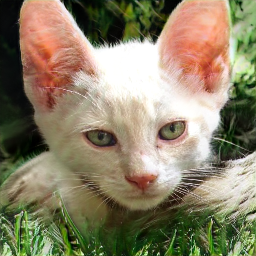}
    \includegraphics[width=0.16\linewidth]{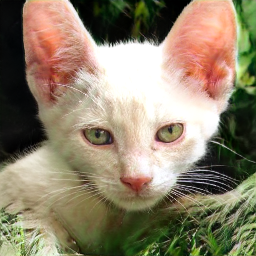}
    \includegraphics[width=0.16\linewidth]{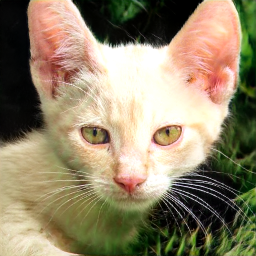}
    \includegraphics[width=0.16\linewidth]{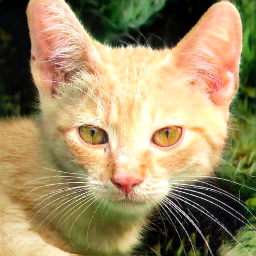}
    }
    \subfigure[Random seed 1469]{\includegraphics[width=0.16\linewidth]{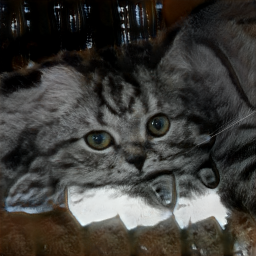}
    \includegraphics[width=0.16\linewidth]{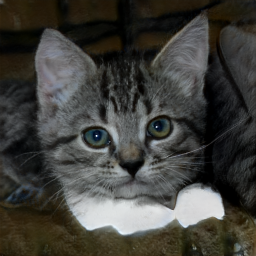}
    \includegraphics[width=0.16\linewidth]{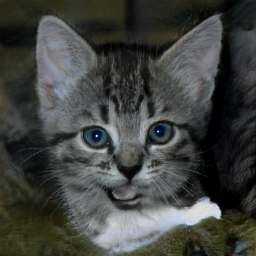}
    \includegraphics[width=0.16\linewidth]{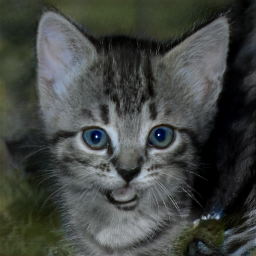}
    \includegraphics[width=0.16\linewidth]{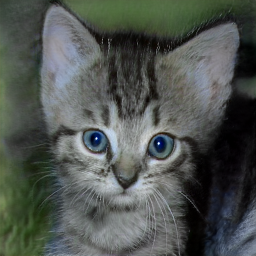}
    \includegraphics[width=0.16\linewidth]{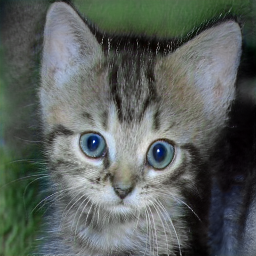}
    }
    \subfigure[Random seed 1645]{\includegraphics[width=0.16\linewidth]{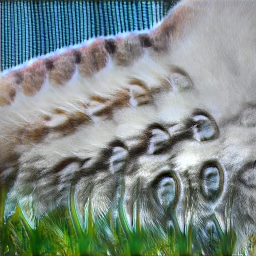}
    \includegraphics[width=0.16\linewidth]{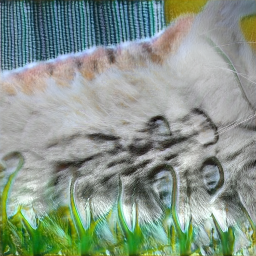}
    \includegraphics[width=0.16\linewidth]{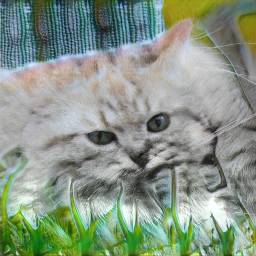}
    \includegraphics[width=0.16\linewidth]{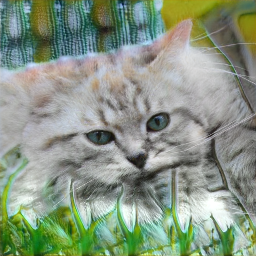}
    \includegraphics[width=0.16\linewidth]{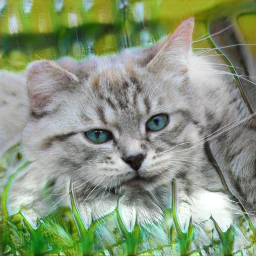}
    \includegraphics[width=0.16\linewidth]{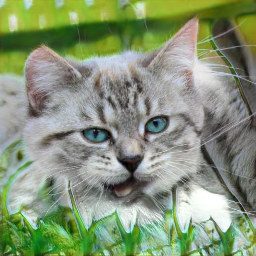}
    }
    \caption{More results of refining StyleGAN2+ADA pre-trained model on AFHQ Cat ($512\times512$). }
    \label{fig:more_cat}
\end{figure*}

\begin{figure*}[h!]
    \centering
    \subfigure[Random seed 234]{\includegraphics[width=0.16\linewidth]{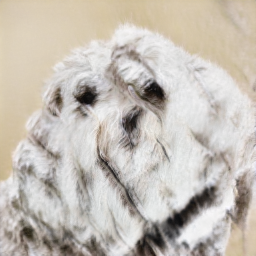}
    \includegraphics[width=0.16\linewidth]{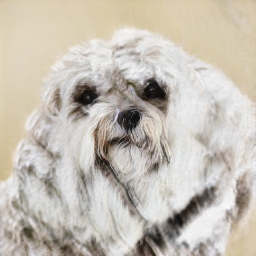}
    \includegraphics[width=0.16\linewidth]{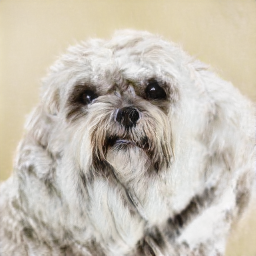}
    \includegraphics[width=0.16\linewidth]{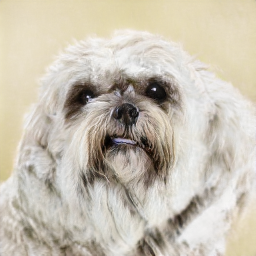}
    \includegraphics[width=0.16\linewidth]{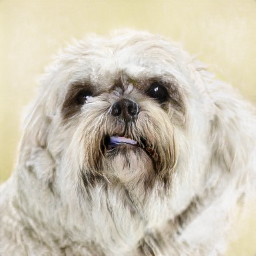}
    \includegraphics[width=0.16\linewidth]{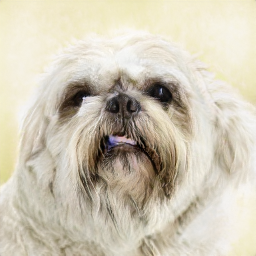}
    }
    \subfigure[Random seed 293]{\includegraphics[width=0.16\linewidth]{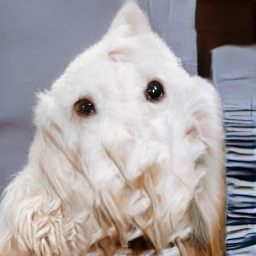}
    \includegraphics[width=0.16\linewidth]{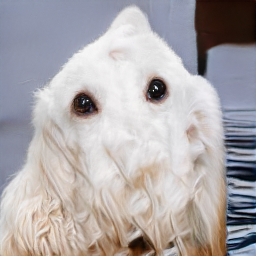}
    \includegraphics[width=0.16\linewidth]{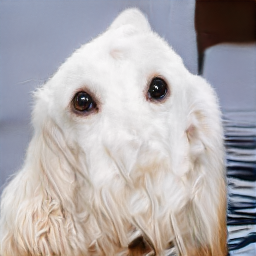}
    \includegraphics[width=0.16\linewidth]{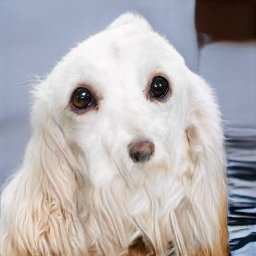}
    \includegraphics[width=0.16\linewidth]{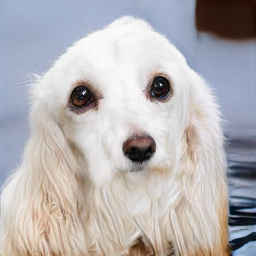}
    \includegraphics[width=0.16\linewidth]{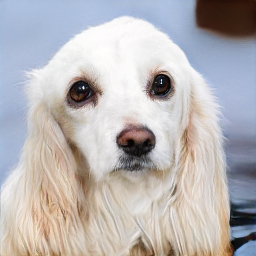}
    }
    \subfigure[Random seed 338]{\includegraphics[width=0.16\linewidth]{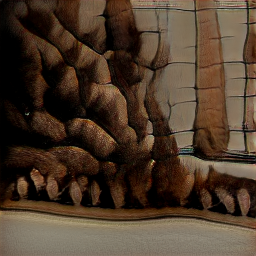}
    \includegraphics[width=0.16\linewidth]{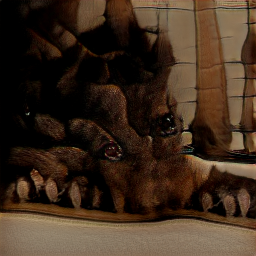}
    \includegraphics[width=0.16\linewidth]{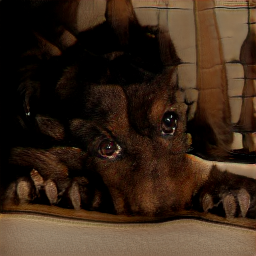}
    \includegraphics[width=0.16\linewidth]{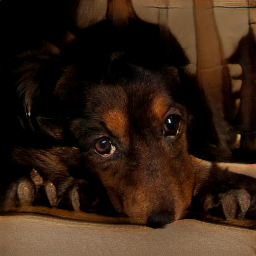}
    \includegraphics[width=0.16\linewidth]{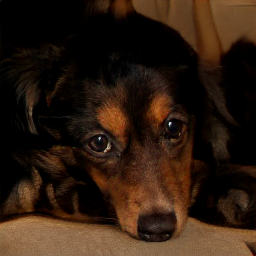}
    \includegraphics[width=0.16\linewidth]{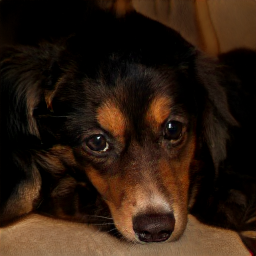}
    }
    \subfigure[Random seed 353]{\includegraphics[width=0.16\linewidth]{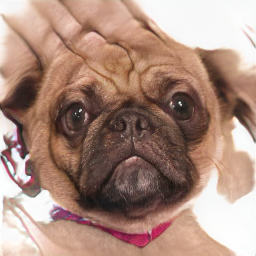}
    \includegraphics[width=0.16\linewidth]{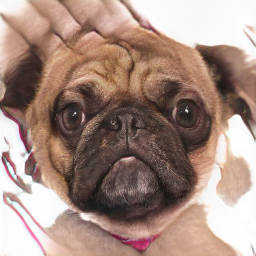}
    \includegraphics[width=0.16\linewidth]{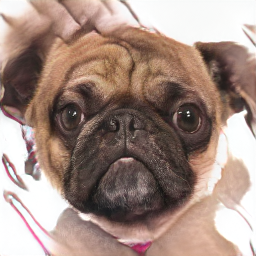}
    \includegraphics[width=0.16\linewidth]{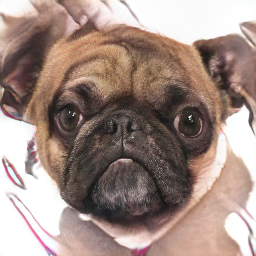}
    \includegraphics[width=0.16\linewidth]{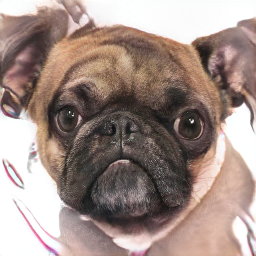}
    \includegraphics[width=0.16\linewidth]{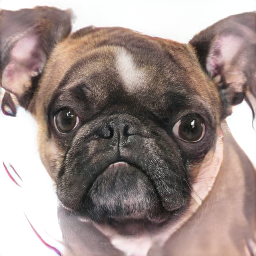}
    }
    \subfigure[Random seed 496]{\includegraphics[width=0.16\linewidth]{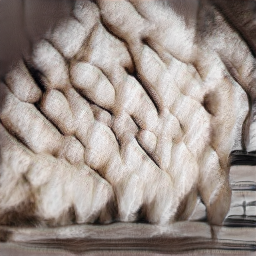}
    \includegraphics[width=0.16\linewidth]{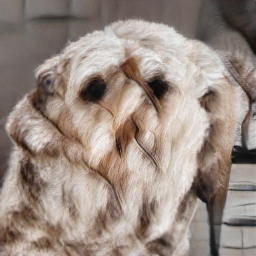}
    \includegraphics[width=0.16\linewidth]{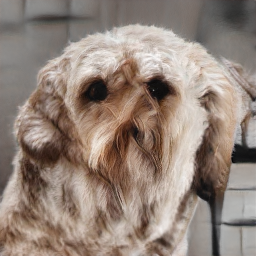}
    \includegraphics[width=0.16\linewidth]{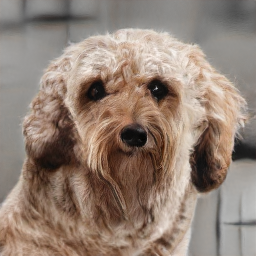}
    \includegraphics[width=0.16\linewidth]{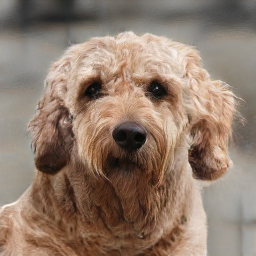}
    \includegraphics[width=0.16\linewidth]{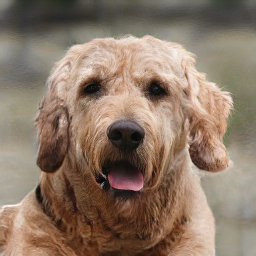}
    }
    \caption{More results of refining StyleGAN2+ADA pre-trained model on AFHQ Dog ($512\times512$). }
    \label{fig:more_dog}
\end{figure*}

\clearpage
\begin{figure*}[h!]
    \centering
    \subfigure[Random seed 541]{\includegraphics[width=0.16\linewidth]{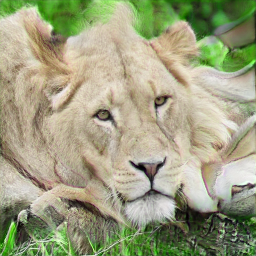}
    \includegraphics[width=0.16\linewidth]{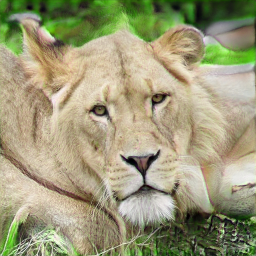}
    \includegraphics[width=0.16\linewidth]{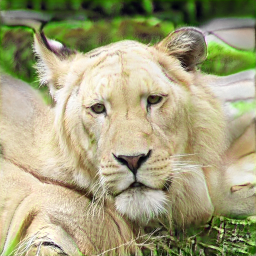}
    \includegraphics[width=0.16\linewidth]{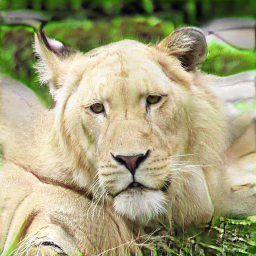}
    \includegraphics[width=0.16\linewidth]{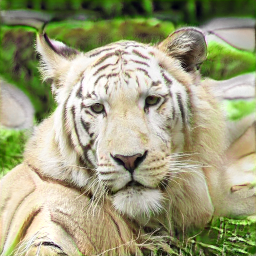}
    \includegraphics[width=0.16\linewidth]{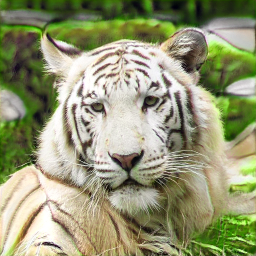}
    }
    \subfigure[Random seed 829]{\includegraphics[width=0.16\linewidth]{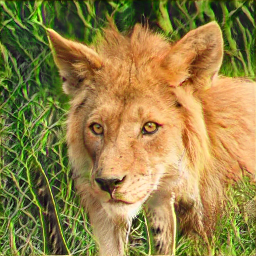}
    \includegraphics[width=0.16\linewidth]{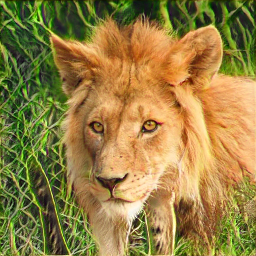}
    \includegraphics[width=0.16\linewidth]{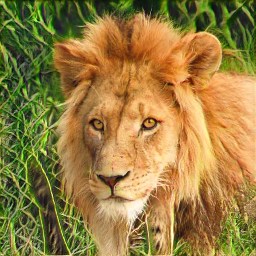}
    \includegraphics[width=0.16\linewidth]{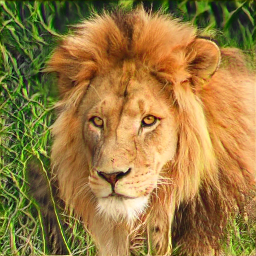}
    \includegraphics[width=0.16\linewidth]{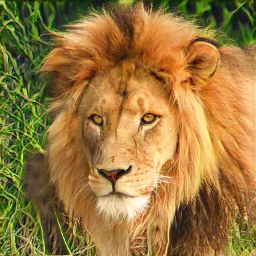}
    \includegraphics[width=0.16\linewidth]{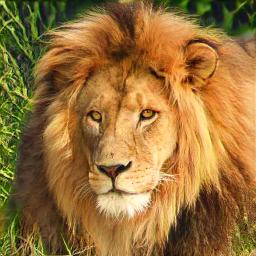}
    }
    \caption{Results of refining StyleGAN2+ADA pre-trained model on AFHQ Wild ($512\times512$). }
    \label{fig:more_wild}
\end{figure*}

\section{More Results on Image Generation}
\begin{figure}[h!]
    \centering
    \subfigure[CIFAR-10 ($32\times32$)]{\includegraphics[width=0.23\linewidth]{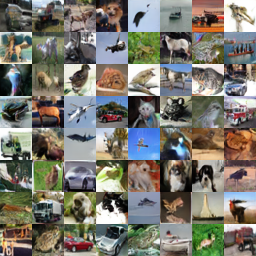}}
    \subfigure[STL-10 ($48\times48$)]{\includegraphics[width=0.23\linewidth]{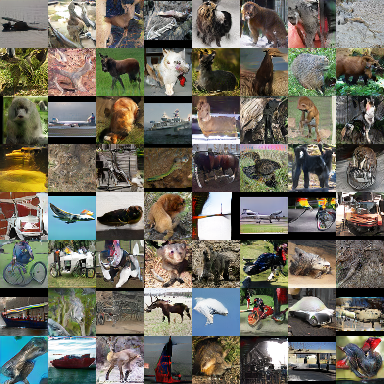}}
    \subfigure[AFHQ Cat ($512\times512$)]{\includegraphics[width=0.23\linewidth]{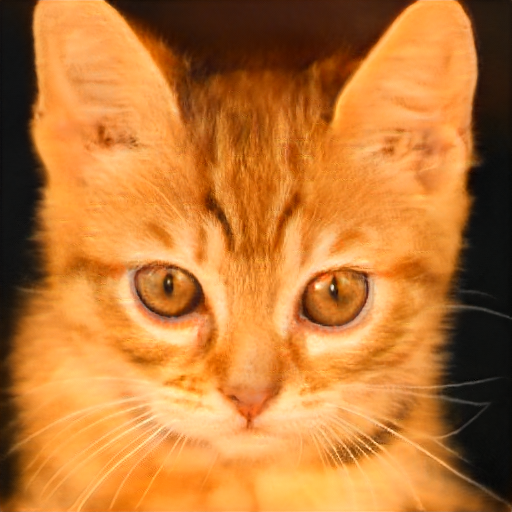}}
    \subfigure[AFHQ Dog ($512\times512$)]{\includegraphics[width=0.23\linewidth]{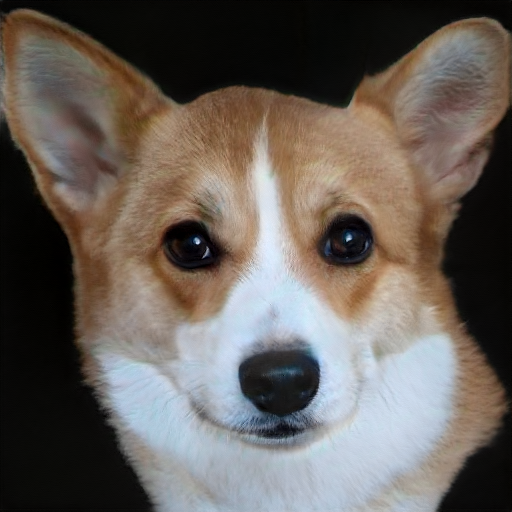}}
    \caption{Generated images using the proposed method.}
    \label{fig:image_generation}
\end{figure}

\begin{table}[h!]
    \caption{FID results of the proposed method on STL-10.}
    \label{tab:different_sampling_cnn_stl10}
    \centering
    \begin{sc}
    \begin{adjustbox}{scale=0.65}
    \begin{tabular}{lccccc}
        \toprule
         \multirow{2}{*}{Steps} &  \multicolumn{5}{c}{Functional}\\
         & Langevin & KL &  RKL & JS & SH \\
         \midrule 
          \multicolumn{6}{c}{Trained without morphing} \\
          0 &  \multicolumn{5}{c}{$33.50 \pm 0.13$} \\
         10 & $27.72 \pm 0.16$ & $24.20 \pm 0.08$ & $28.01 \pm 0.10$ & $24.49 \pm 0.09$ & $25.01 \pm 0.06$\\
         20 & $27.28 \pm 0.05$ & $23.18 \pm 0.04$ & $26.41 \pm 0.08$ & $23.33 \pm 0.08$ & $23.29 \pm 0.04$\\
         30 & $27.80 \pm 0.09$ & $22.36 \pm 0.07$ & $24.61 \pm 0.06$ & $22.36 \pm 0.07$ & $22.50 \pm 0.05$\\
         40 & $27.28 \pm 0.08$ & $22.82 \pm 0.05$ & $24.32 \pm 0.12$ & $\mathbf{22.15 \pm 0.07}$ & $22.27 \pm 0.12$\\
         \midrule
         \multicolumn{6}{c}{Trained with 5-step morphing} \\
         0 &  \multicolumn{5}{c}{$32.82 \pm 0.07$} \\
         10 & $26.26 \pm 0.09$ & $ 22.22 \pm 0.09 $ & $24.15 \pm 0.11$ & $23.81 \pm 0.09$ & $23.37 \pm 0.08$\\
         20 & $26.61 \pm 0.07$ & $ 21.52 \pm 0.13 $ & $23.06 \pm 0.10$ & $22.56 \pm 0.06$ & $22.17 \pm 0.06$\\
         30 & $27.16 \pm 0.12$ & $\mathbf{21.40 \pm 0.05}$ & $22.69\pm 0.06$ & $21.77 \pm 0.07$ & $21.69  \pm 0.07$\\
         40 & $26.76 \pm 0.15$ & $21.49 \pm 0.07$ & $22.65 \pm 0.10$ & $21.91 \pm 0.14$ & $21.89 \pm 0.03$\\
         \bottomrule
    \end{tabular}
    \end{adjustbox}
    \end{sc}
\end{table}

\clearpage
\section{More Results of Image Translation}

\begin{figure*}[h!]
    \centering
    \includegraphics[width=0.9\linewidth]{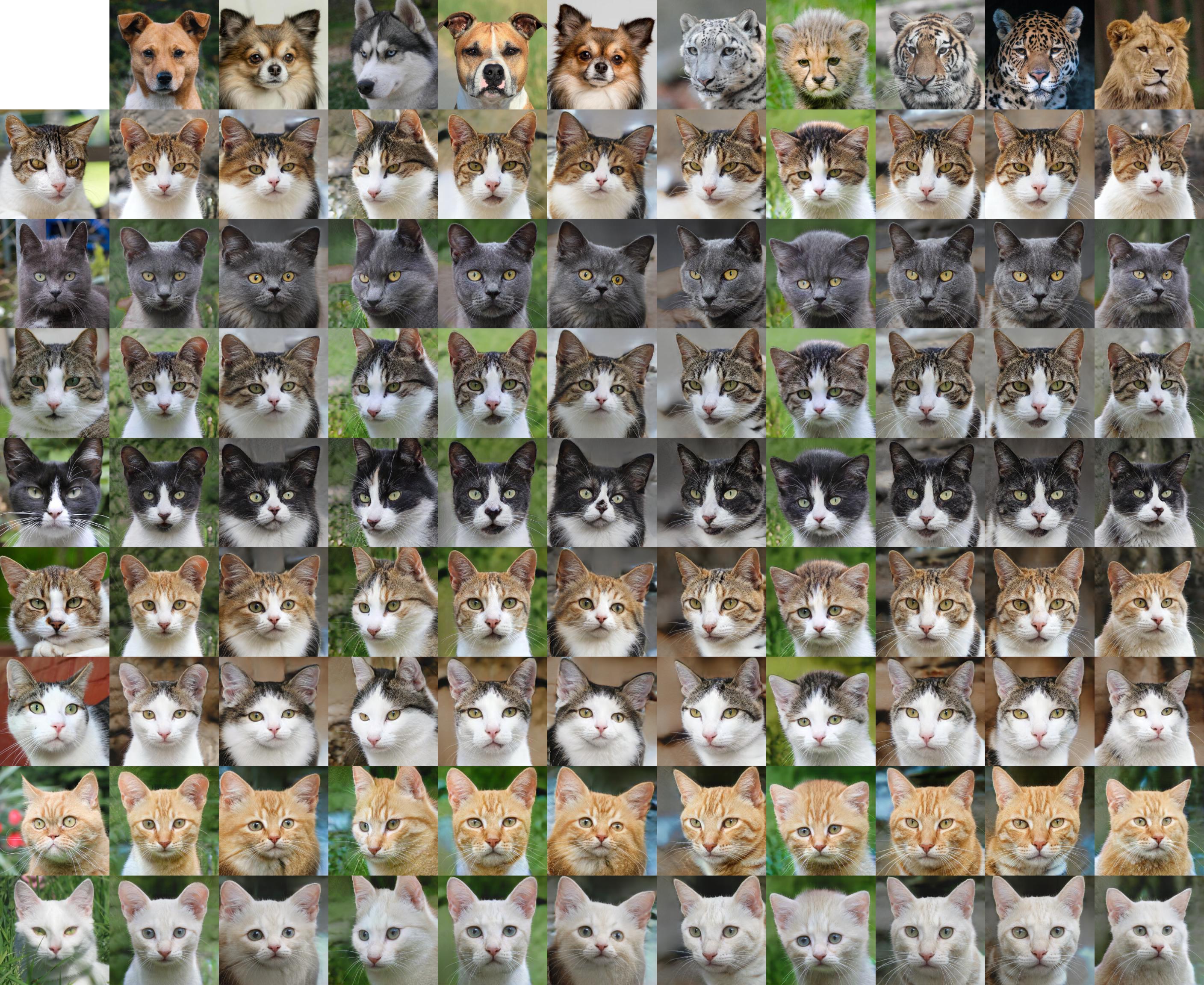}
    \caption{More results of image translation on AFHQ dataset. The first row images are source images, the first column images are reference images.}
\end{figure*}

\begin{figure*}[h!]
    \centering
    \includegraphics[width=0.9\linewidth]{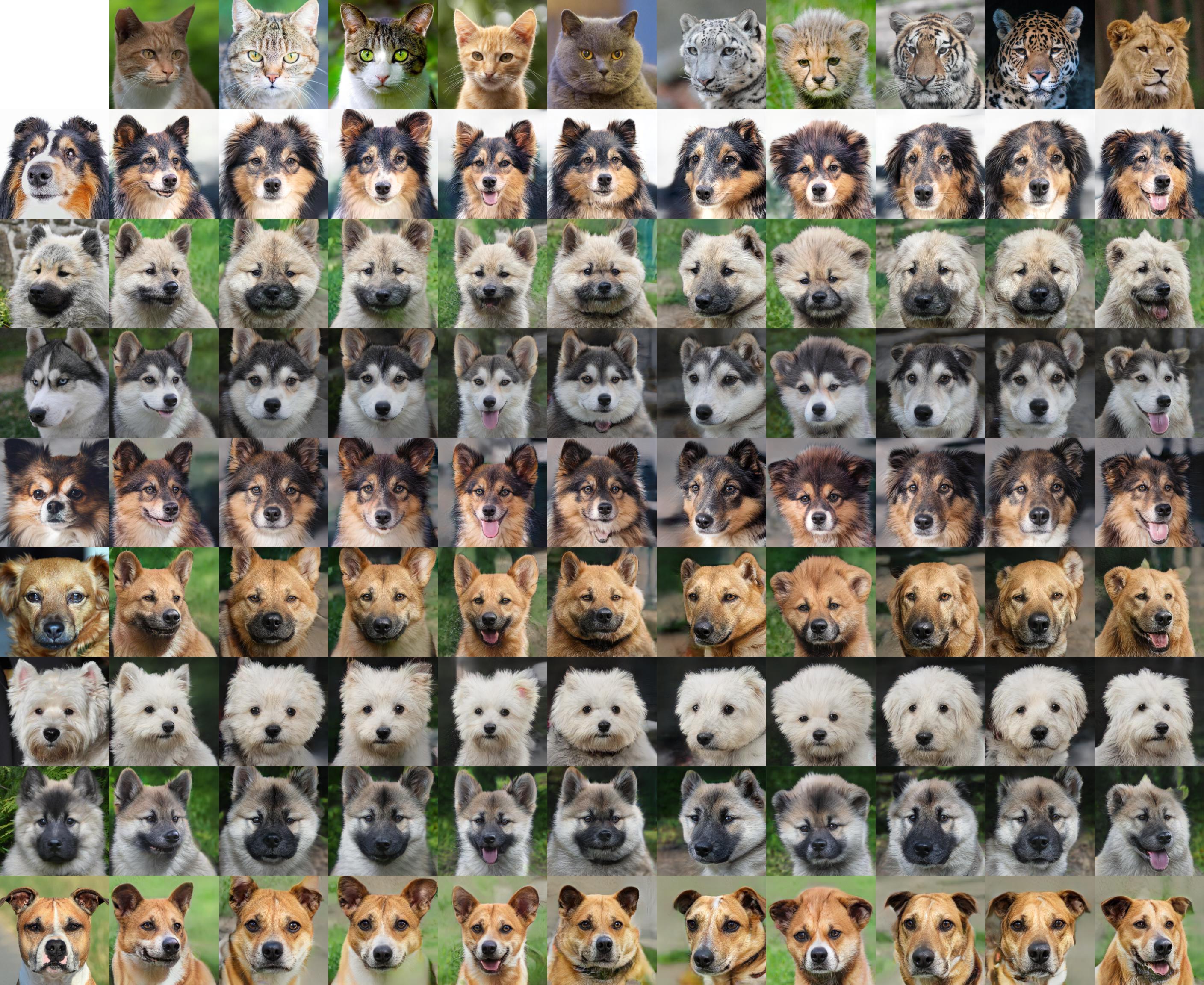}
    \caption{More results of image translation on AFHQ dataset. The first row images are source images, the first column images are reference images.}
\end{figure*}
\begin{figure*}[h!]
    \centering
    \includegraphics[width=0.9\linewidth]{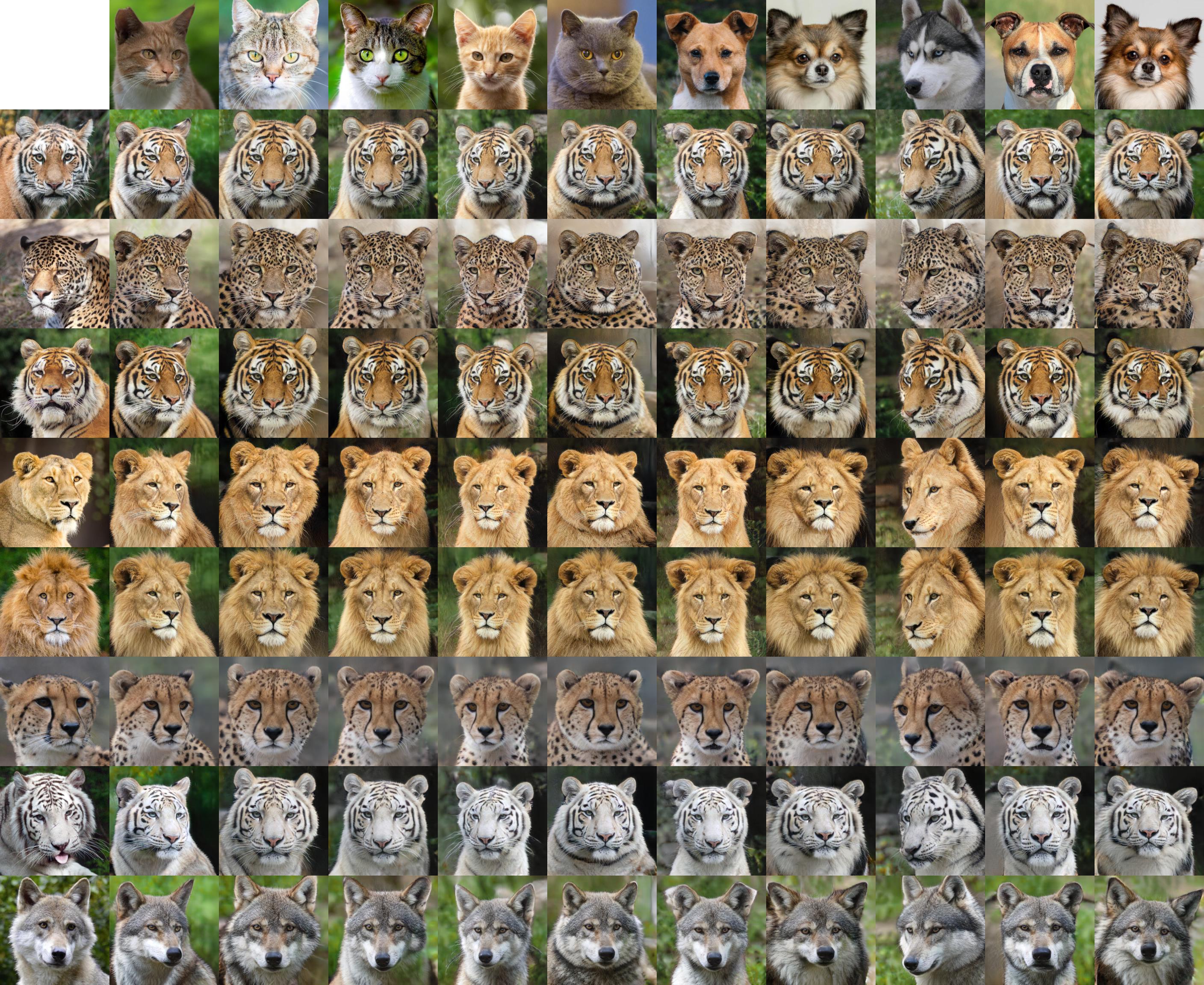}
    \caption{More results of image translation on AFHQ dataset. The first row images are source images, the first column images are reference images.}
\end{figure*}
\begin{figure*}[h!]
    \centering
    \includegraphics[width=0.9\linewidth]{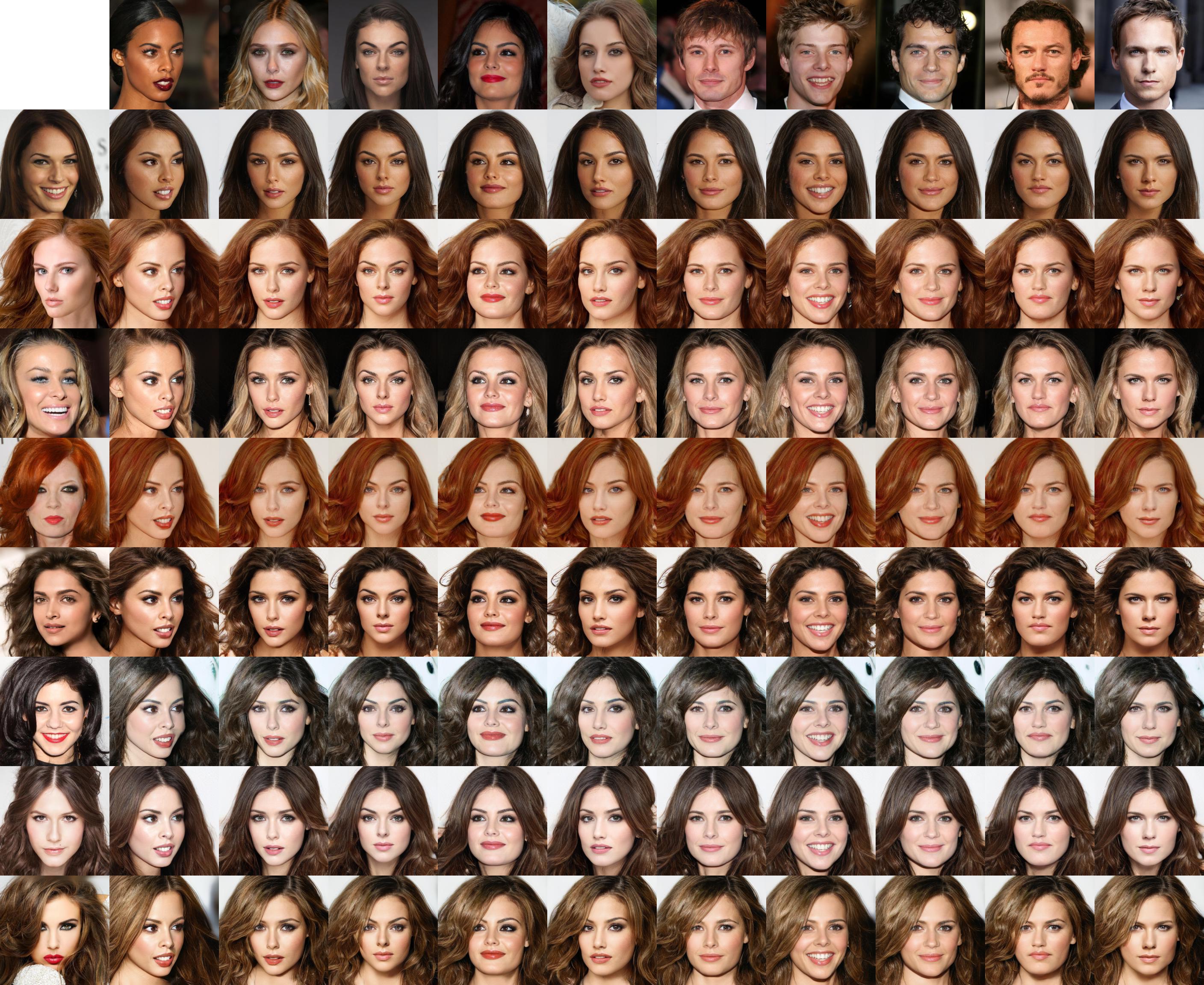}
    \caption{More results of image translation on CelebA-HQ dataset. The first row images are source images, the first column images are reference images.}
\end{figure*}
\begin{figure*}[h!]
    \centering
    \includegraphics[width=0.9\linewidth]{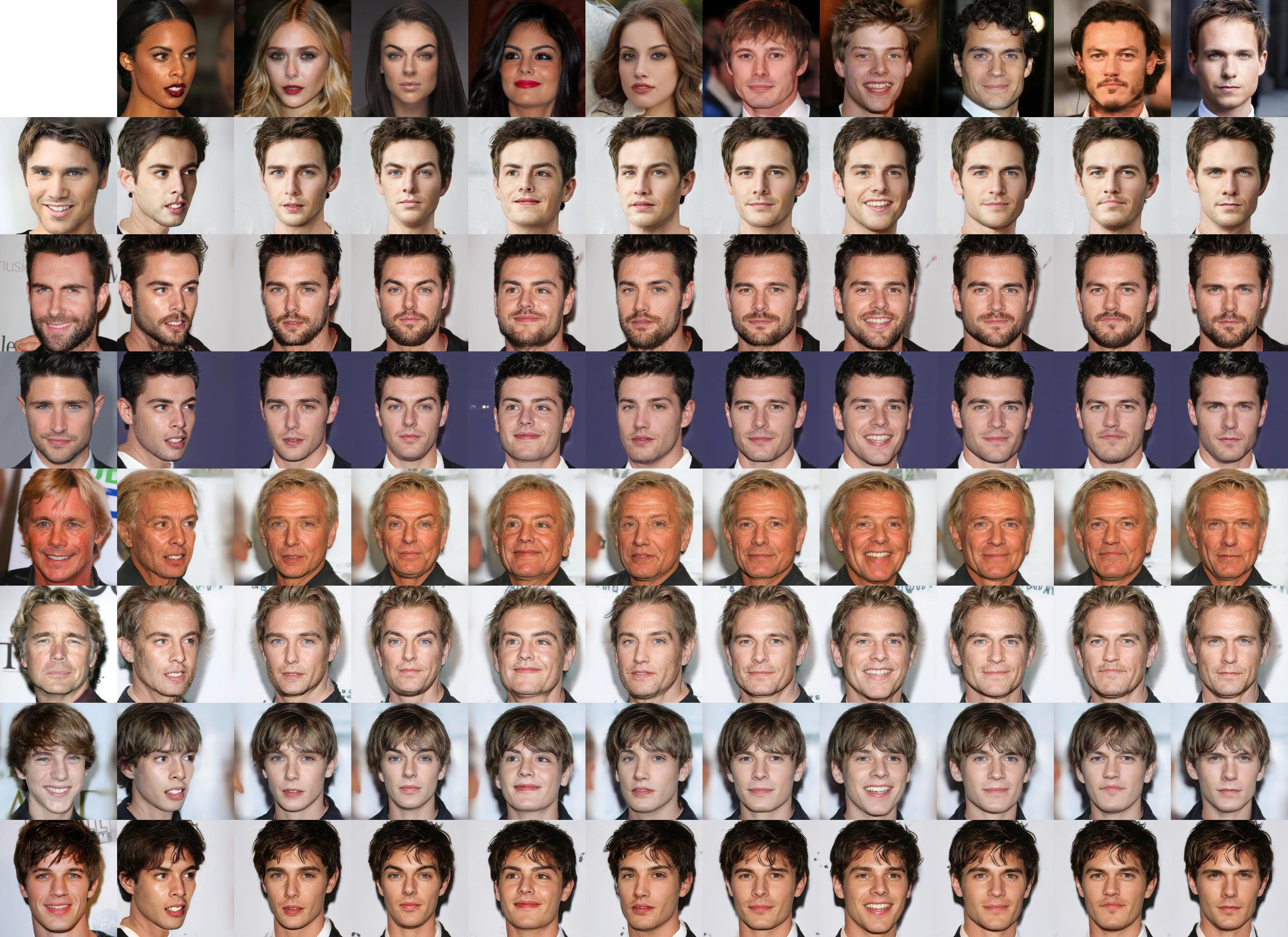}
    \caption{More results of image translation on CelebA-HQ dataset. The first row images are source images, the first column images are reference images.}
\end{figure*}


\end{document}